\documentclass{article}

     \PassOptionsToPackage{numbers, compress}{natbib}
\usepackage[final]{neurips_2021}

\usepackage[utf8]{inputenc} % allow utf-8 input
\usepackage[T1]{fontenc}    % use 8-bit T1 fonts
\usepackage{hyperref}       % hyperlinks
\usepackage{url}            % simple URL typesetting
\usepackage{booktabs}       % professional-quality tables
\usepackage{amsfonts}       % blackboard math symbols
\usepackage{nicefrac}       % compact symbols for 1/2, etc.
\usepackage{microtype}      % microtypography
\usepackage{xcolor}         % colors
\usepackage{wrapfig}
\DeclareMathAlphabet\mathbfcal{OMS}{cmsy}{b}{n}
\def\0{{\bf 0}}
\def\1{{\bf 1}}

\def\mmE{{\mathbb E}}

\newtheorem{thm}{Theorem}

\newenvironment*{proof}{\textbf{Proof}\quad}{\hfill $\square$\par}

\def\eg{\emph{e.g.}} 
\def\ie{\emph{i.e.}}

\def\wrt{{w.r.t.}} 
 
\usepackage{tabu} 

\usepackage{microtype}
\usepackage{graphicx}
\usepackage{subfigure} 
\usepackage{bm}
\usepackage{comment}
\usepackage{amsfonts}
\usepackage{amsmath}
\usepackage{balance}
\usepackage{color}
\usepackage{listings} 
\usepackage{xcolor} 
\usepackage{arydshln}
\usepackage{cases} 
\usepackage{booktabs}
\usepackage{algorithm}
\usepackage{algorithmic} 
\usepackage{threeparttable}
\usepackage{multicol,multirow}   
\def\ie{\emph{i.e., }}
\def\eg{\emph{e.g., }}
\def\wrt{\emph{w.r.t. }}

\usepackage{hyperref}
\renewcommand{\algorithmicensure}{\textbf{Initialize:}}

\title{Unleashing the Power of Contrastive Self-Supervised Visual Models via Contrast-Regularized Fine-Tuning}

\author{%
  Yifan Zhang$^1$\thanks{Corresponding to: Yifan Zhang <yifan.zhang@u.nus.edu>} \quad  Bryan Hooi$^1$ \quad Dapeng Hu$^1$ \quad Jian Liang$^2$  \quad Jiashi Feng$^{3}$ \\ 
  $^1$National University of Singapore \quad $^2$Chinese Academy of Sciences \quad $^3$SEA AI Lab 
}
\begin{document}
\maketitle

\begin{abstract}
Contrastive self-supervised learning (CSL) has attracted increasing attention for model pre-training via  unlabeled data. The resulted CSL models provide instance-discriminative visual features that are uniformly scattered in the feature space.  During deployment, the common practice is to directly fine-tune CSL models with cross-entropy, which however may not be  the best strategy in practice. Although cross-entropy tends to separate inter-class features, the resulting models still have limited capability for reducing intra-class feature scattering that exists in CSL models. In this paper, we investigate whether applying contrastive learning to fine-tuning would bring further benefits, and analytically find that optimizing the contrastive loss benefits both discriminative representation learning and model optimization during fine-tuning. Inspired by these findings, we propose Contrast-regularized tuning (Core-tuning), a new approach for fine-tuning CSL models. Instead of simply adding the contrastive loss to the objective of fine-tuning, Core-tuning further applies a novel hard pair mining strategy for more effective contrastive fine-tuning, as well as smoothing the decision boundary to better exploit  the learned discriminative feature space. Extensive experiments on image classification and semantic segmentation verify the effectiveness of Core-tuning.
\end{abstract}   
\vspace{-0.1in}

\section{Introduction}\label{Introduction}
Pre-training a deep neural network on a large database and then fine-tuning it on downstream tasks has been a popular training scheme. Recently,  contrastive self-supervised learning (CSL) has attracted increasing attention on model pre-training, since it does not rely on any hand-crafted annotations but even achieves more promising performance than supervised pre-training on downstream tasks  ~\cite{chen2020simple,chen2020big,he2020momentum,hu2021well,tian2020makes}. 
Specifically, CSL leverages unlabeled data to train visual models via contrastive learning, which maximizes the feature similarity  for two augmentations of the same instance and minimizes the feature  similarity of two instances~\cite{wu2018unsupervised}.
The learned models  provide instance-discriminative visual representations that are uniformly scattered in the feature space~\cite{wang2020understanding}.

Although there have been  substantial CSL studies on model pre-training~\cite{kalantidis2020hard,tian2020makes}, few have explored the fine-tuning process. The common practice is to directly fine-tune CSL models  with the cross-entropy loss~\cite{chen2020simple,ericsson2021well,he2020momentum}. However,  we empirically  (cf.~Table~\ref{exp_classification}) find  that different fine-tuning methods significantly influence the model performance on downstream tasks, and fine-tuning with only cross-entropy is not the optimal strategy.
Intuitively, although cross-entropy tends to learn separable features among classes, the resulting model is still limited in its capability for reducing intra-class feature scattering~\cite{liu2016large,wen2016discriminative} that exists in CSL models. Meanwhile,
most existing fine-tuning methods~\cite{li2019delta,li2018explicit}  are devised for supervised pre-trained models and tend to enforce regularizers to prevent the fine-tuned models changing too much from the pre-trained ones.
However, they suffer from the issue of negative transfer~\cite{chen2019catastrophic}, since downstream tasks are often different from the pre-training contrastive task.  
In this sense, how to  fine-tune CSL models  remains an important yet under-explored question.

\begin{figure*}[t]
 \vspace{-0.1in} 
 \begin{minipage}{1\linewidth}
 \centerline{\includegraphics[width=9.5cm]{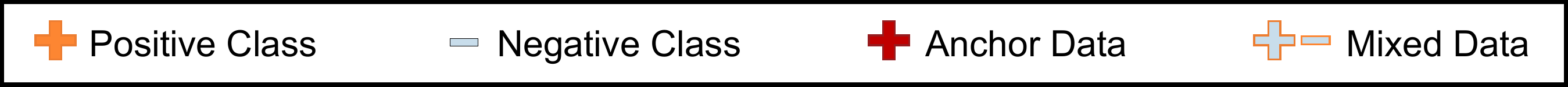}} 
 \end{minipage}  \vspace{0.01in}
 \vfill  
 \begin{minipage}{1\linewidth}
 \hspace{0.05in}\centerline{\includegraphics[width=13.5cm]{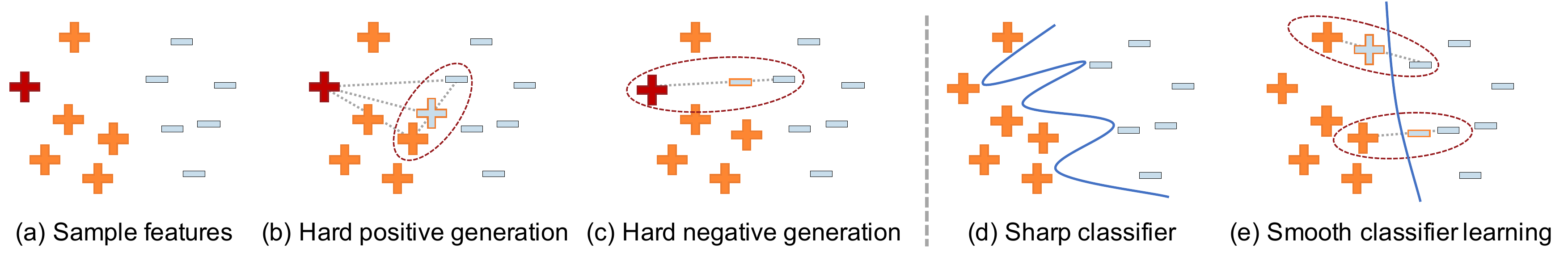}}  
 \end{minipage}% \vspace{-0.15in}
 \caption{Illustration of two challenges in contrastive fine-tuning. (1) How to mine hard sample pairs for more effective contrastive fine-tuning. As shown in (a), the majority of sample pairs are easy-to-contrast, which may induce negligible contrastive loss gradients that contribute little to learning discriminative representations.  
 (2) How to improve the generalizability of the model. As shown in (d), the classifier simply trained  with cross-entropy is often sharp and near training data, leading to limited generalization performance.}
 %To address this,  Core-tuning develops a new hard pair mining strategy for more effective contrastive fine-tuning.%To enhance generalizability, Core-tuning further smooths the decision boundary to better exploit the learned discriminative feature space.
 \vspace{-0.2in} 
 \label{challenges_motivations} 
\end{figure*}   
  
Considering that optimizing the unsupervised contrastive loss during pre-training yields models with instance-level discriminative power, we investigate whether applying contrastive learning to fine-tuning would bring further benefits.
To answer this, we analyze the contrastive loss during fine-tuning (cf.~Section~\ref{contrastive loss}) and find that it offers two benefits.
First, integrating the contrastive loss into cross-entropy can provide an additional regularization effect, as compared to cross-entropy based fine-tuning, for discriminative representation learning.
Such an effect encourages the model to learn a low-entropy feature cluster for each class  (\ie high intra-class compactness)  and a high-entropy feature space  (\ie large inter-class separation degree).
%See feature visualization on CIFAR10 in Figure~\ref{visualization} for an illustration.
Second, optimizing the contrastive loss will  minimize the infimum of the cross-entropy loss over training data, which can provide  an additional optimization effect for model fine-tuning. Based on the optimization effectiveness as well as the regularization effectiveness on representations, we argue that optimizing the contrastive loss during fine-tuning  can further improve the performance of CSL models on downstream tasks.

Considering the above benefits, a natural idea is to  directly add the contrastive loss to the objective for fine-tuning, \eg one recent study~\cite{gunel2020supervised} simply uses contrastive learning to fine-tune language models. However, such a method cannot take full advantage of contrastive learning, since it ignores an important challenge in contrastive fine-tuning. That is, contrastive learning highly relies on positive/negative sample pairs, but the majority of sample features are easy-to-contrast (cf.~Figure~\ref{challenges_motivations} (a))~\cite{harwood2017smart,wu2017sampling} and may produce negligible contrastive loss gradients. Ignoring this makes the method~\cite{gunel2020supervised} fail to learn  more discriminative features via contrastive learning and thus cannot fine-tune CSL models well.

In this paper, to better fine-tune CSL models and enhance their performance on downstream tasks, we propose a contrast-regularized tuning approach (termed Core-tuning), based on a novel hard pair mining strategy. % for more effective contrastive learning
Specifically, Core-tuning  generates both hard positive and hard negative pairs for each anchor data  via a new hardness-directed mixup strategy (cf.~Figure~\ref{challenges_motivations} (b-c)). Here, hard positives indicate the positive pairs far away from the anchor, while hard negatives are the negative pairs close to the anchor. Meanwhile, since hard pairs are more informative for contrastive learning~\cite{harwood2017smart},  Core-tuning further assigns higher importance weights to hard positive pairs  based on a new focal contrastive loss. 
In this way, the resulting model   is able to learn a more discriminative feature space by contrastive fine-tuning.
Following that, we also explore how to better exploit the learned discriminative feature space in  Core-tuning. % also seeks to   better% by smoothing the decision boundary.
Previous work has found that the decision boundary simply trained with cross-entropy is often sharp and close to training data~\cite{verma2019manifold}, which may make the classifier fail to exploit the high inter-class separation degree in the discriminative feature space (cf.~Figure~\ref{challenges_motivations} (d)), and also suffer from   limited generalization performance.
To address this, Core-tuning further uses the mixed features to train the classifier, so that the learned decision boundaries can be more smooth and far away from the original training data (cf.~Figure~\ref{challenges_motivations} (e)). 
%Inspired by~\cite{verma2019manifold} in which mixed data are used to help learn smoother decision boundaries
% lead to incorrect yet confident predictions when evaluated on slightly different test samples
%Extensive experiments on image classification and semantic segmentation verify the effectiveness of Core-tuning. 
%We also empirically find that Core-tuning benefits contrastive self-supervised models in terms of  domain generalization and adversarial robustness on downstream tasks.

The key contributions  are threefold. 
1)  To our knowledge, we are among the first to look into the fine-tuning stage of CSL models, which is  an important yet under-explored question. To address this, we propose a novel Core-tuning method.
2) We theoretically analyze the benefits of the supervised contrastive loss on representation learning and model optimization, revealing that it is beneficial to model fine-tuning. 
3) Promising results on image classification and semantic segmentation verify the effectiveness of Core-tuning for improving the fine-tuning performance of CSL models. We also empirically find that Core-tuning benefits CSL models in terms of  domain generalization and adversarial robustness on downstream tasks.
Considering the theoretical guarantee and empirical effectiveness of Core-tuning, we recommend using it as a standard baseline to fine-tune CSL models.
%bh: I think it is better to give a reason here, otherwise the statement feels unjustified. E.g. `Considering the simplicity and general applicability of Core-tuning, we recommend using it as a standard baseline for the fine-tuning of contrastive self-supervised visual models.'  

%We empirically demonstrate that Core-tuning   effectively improve the fine-tuning performance of CSL models on many downstream tasks, including image classification, semantic segmentation. 
 
%\newpage
\section{Related Work}  
\textbf{Contrastive self-supervised learning (CSL).} 
Self-supervised learning is a kind of unsupervised learning method based on self-supervised proxy tasks, \eg  rotation prediction~\cite{gidaris2018unsupervised}, colorization prediction~\cite{larsson2016learning} and clustering~\cite{yan2020clusterfit}.
Recently, CSL has become the most popular self-supervised paradigm, which treats each instance as a category to learn instance-discriminative representations. 
State-of-the-art CSL methods include InsDis~\cite{wu2018unsupervised},   MoCo~\cite{he2020momentum}, SimCLR~\cite{chen2020simple,chen2020big} and InfoMin~\cite{tian2020makes}.
Most CSL studies are devoted to network pre-training, but few have explored the fine-tuning process.

As an effective data augmentation method, mixup~\cite{zhang2020does} has recently been applied to instance augmentation for  CSL~\cite{kalantidis2020hard,kim2020mixco,lee2020mix,shen2020rethinking}. Among these methods, the work~\cite{kalantidis2020hard} uses mixup to generate hard negative pairs for better instance discrimination. However, all these methods focus on unsupervised pre-training and cannot accurately generate hard pairs regarding classes. Comparatively, Core-tuning focuses on the fine-tuning of CSL models and can generate accurate hard positive/negative pairs for each class. Note that the hardness-directed mixup strategy in Core-tuning is different from manifold mixup~\cite{verma2019manifold} that cannot be directly used to generate hard sample pairs.    

\textbf{Pre-training and Fine-tuning.}  
In deep learning, it is a popular scheme to first pre-train a deep neural network on a large database (\eg ImageNet) and then fine-tune it on downstream tasks~\cite{li2018explicit,li2020rifle}. 
Supervised learning is the mainstream method for pre-training~\cite{kornblith2019better}, whereas self-supervised learning is attracting increasing attention since it  does not rely on rich annotations~\cite{chen2020simple,chen2020big}. 
Most existing methods for fine-tuning, like L2-SP~\cite{li2018explicit} and DELTA~\cite{li2019delta}, are devised for supervised pre-trained models and tend to enforce some regularizer to prevent the fine-tuned models changing too much from the pre-trained ones.
However, they may be unsuitable for contrastive self-supervised models, since downstream tasks are often different from the  contrastive pre-training task, leading to negative transfer~\cite{chen2019catastrophic}.  
Very recently, one work~\cite{gunel2020supervised} explored  contrastive learning to   fine-tune language models.
However, it simply add  the contrastive loss to the objective of fine-tuning and cannot theoretically explain why it boosts fine-tuning.  More critically, it ignores the challenge of hard pair mining in contrastive fine-tuning    and thus cannot fine-tune CSL models well.

\section{Effects of Contrastive Loss for Model Fine-tuning}
\label{contrastive loss}
%\JS{add a prologue here to explain the connections of the following sections, the analysis, the new algorithm for sample generation and the smooth classifier training.}  
%\JS{Need to define the setting this paper is discussing, a model is pre-trained with CSL and then fine-tuned with cross-entropy loss or other loss.} %by diverse CSL methods 
We start by analyzing the benefits of the contrastive loss during fine-tuning, which will motivate our new method. Before that, we first define the problem and notations.

\textbf{Problem Definition and Notation.} This paper studies the fine-tuning of contrastive self-supervised visual models that are pre-trained on a large-scale unlabeled  database. 
During fine-tuning, let $\{(x_i,y_i)\}_{i=1}^n$ denote the target task dataset with $n$ samples, where $x_i$ is an instance with one-hot label $y_i \small{\in} \mathbb{R}^K$ and $K$ denotes the number of classes.   
The neural network model is denoted  by $G$, which consists of a pre-trained feature encoder $G_e$ and a new predictor  $G_y$ specific to the target task.
Based on the network, we extract visual representations by $z_i\small{=}G_e(x_i)$ and make a prediction by $\hat{y}_i\small{=}G_y(z_i)$. Such a contrastive self-supervised model is generally fine-tuned with the cross-entropy loss~\cite{ericsson2021well,he2020momentum}.
  
Following~\cite{boudiaf2020unifying},  we define the random variables of samples  and labels as $X$  and $Y$, and  those of   embeddings and predictions as $Z|X\small{\sim}G_e(X)$ and $\hat{Y}|Z\small{\sim}G_y(Z)$, respectively. Moreover, let $p_Y$ be  the distribution of $Y$,  $p_{(Y,Z)}$ be the joint distribution of $Y$ and $Z$, and  $p_{Y|Z}$ be the conditional distribution of $Y$ given $Z$. We define the entropy of $Y$ as $\mathcal{H}(Y)\small{:=}\mmE_{p_Y}[-\log p_Y(Y)]$ and the conditional entropy of $Y$ given $Z$ as $\mathcal{H}(Y|Z)\small{:=}\mmE_{p_{(Y,Z)}}[-\log p_{Y|Z}(Y|Z)]$. 
Besides, we define the cross-entropy (CE) between $Y$ and $\hat{Y}$    by $\mathcal{H}(Y;\hat{Y})\small{:=}\mmE_{p_Y}[\small{-}\log p_{\hat{Y}}(Y)]$ and the conditional CE given $Z$  by $\mathcal{H}(Y;\hat{Y}|Z)\small{:=}\mmE_{p_{(Y,Z)}}[\small{-}\log p_{\hat{Y}|Z}(Y|Z)]$. 
Before our analysis, we first revisit contrastive loss.
%Lastly, the mutual information between $Z$ and $Y$ is defined as $\mathcal{I}(Z;Y)\small{:=}\mathcal{H}(Y)-\mathcal{H}(Y|Z)$. 

\textbf{Contrastive loss.}  We use the supervised contrastive loss~\cite{khosla2020supervised} for fine-tuning, which is a variant of InfoNCE~\cite{oord2018representation}. Specifically, given a sample feature $z_i$ as anchor, the contrastive loss takes the features from the same class to the anchor as positive pairs  and those from different classes as negative pairs. Assuming features are $\ell_2$-normalized, the contrastive loss is computed by:
\begin{align}\label{eq1}
    \mathcal{L}_{con} \small{=-}\frac{1}{n}\sum_{i=1}^n \frac{1}{|P_i|}\sum_{z_j\in P_i} \log \frac{e^{(z_i^{\top} z_j/\tau)}}{\sum_{z_k\small{\in} A_i}e^{(z_i^{\top} z_k/\tau)}},
\end{align}
where $\tau$ is a temperature factor, while $P_i$ and $A_i$  denote the positive pair set  and the full pair set of the anchor $z_i$, respectively. 
We next analyze the contrastive loss and find it has two beneficial effects.

%\JS{is this specific for fine-tuning? if you are saying something for fine-tuning, at the beginning need to make the setting clear that you are considering applying supervised contrastive learning for model fine-tuning. }

\subsection{Regularization Effect of Contrastive Loss}
\label{Sec:regularization1}
We first show the contrastive loss has regularization effectiveness on representation learning based on the following theorem.
\begin{thm}\label{thm0}
Assuming the features are $\ell_2$-normalized and the classes are balanced with equal data number, minimizing the contrastive loss is equivalent to  minimizing the class-conditional entropy $\mathcal{H}(Z|Y)$ and maximizing the feature entropy $\mathcal{H}(Z)$: 
\begin{align}
  \mathcal{L}_{con} ~    \propto ~~    \mathcal{H}(Z|Y) ~ - ~\mathcal{H}(Z)  \nonumber
\end{align}   
\end{thm}

  \begin{wrapfigure}{r}{0.55\textwidth}
  \begin{center}  
  \vspace{-17pt}
  \begin{minipage}{0.49\linewidth}
 \centerline{\hspace{-0.15in}\includegraphics[width=2.8cm,clip,trim={4.3cm 7.7cm 2.5cm 5.5cm}]{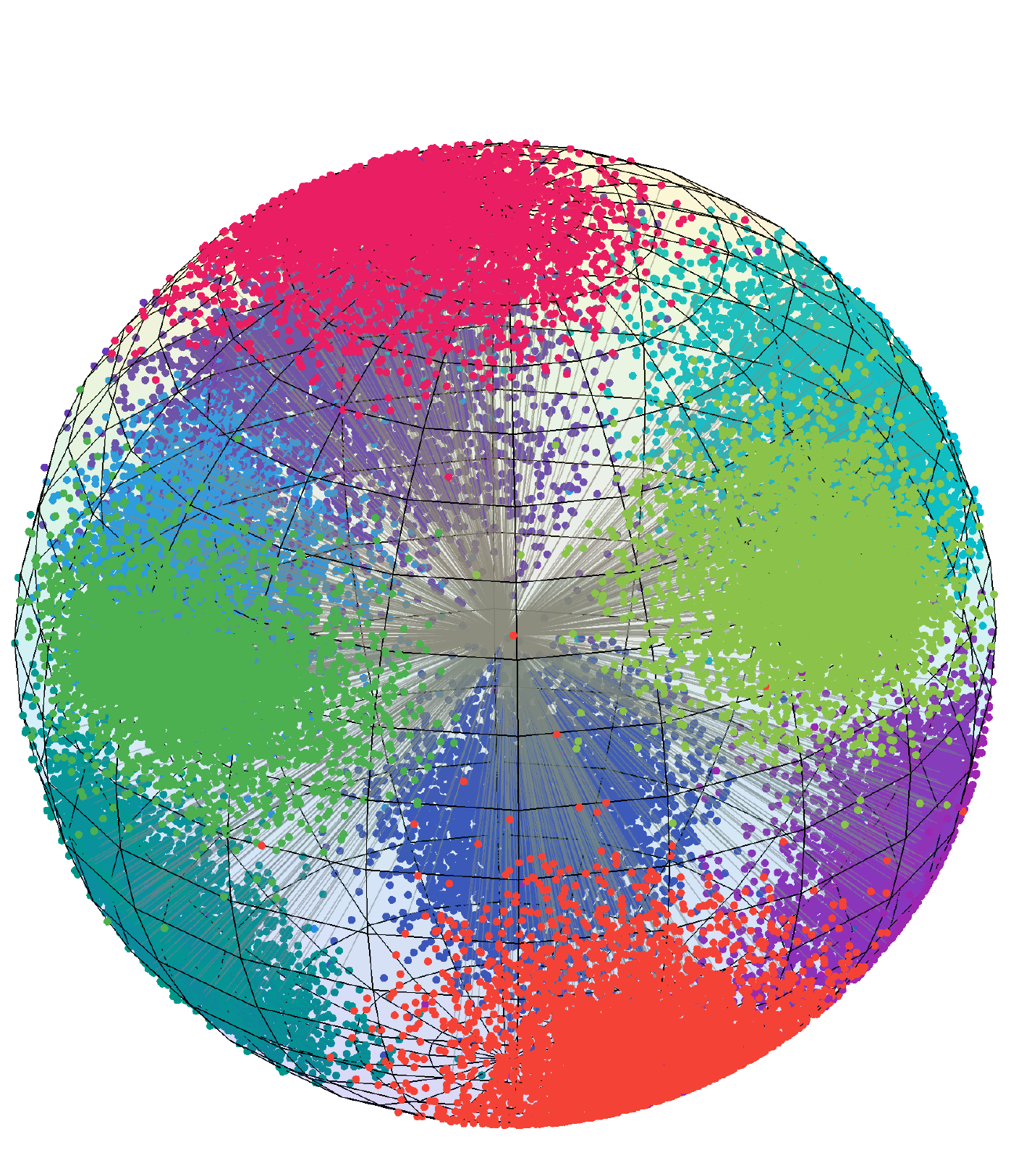}}
 %\vskip -0.25in
  \centerline{\hspace{-0.15in}(a) Training with $\mathcal{L}_{ce}$}
 \end{minipage}
 \hfill 
 \begin{minipage}{0.49\linewidth}
 \centerline{\hspace{-0.2in}\includegraphics[width=2.8cm,clip,trim={4.3cm 7.7cm 2.5cm 5.5cm}]{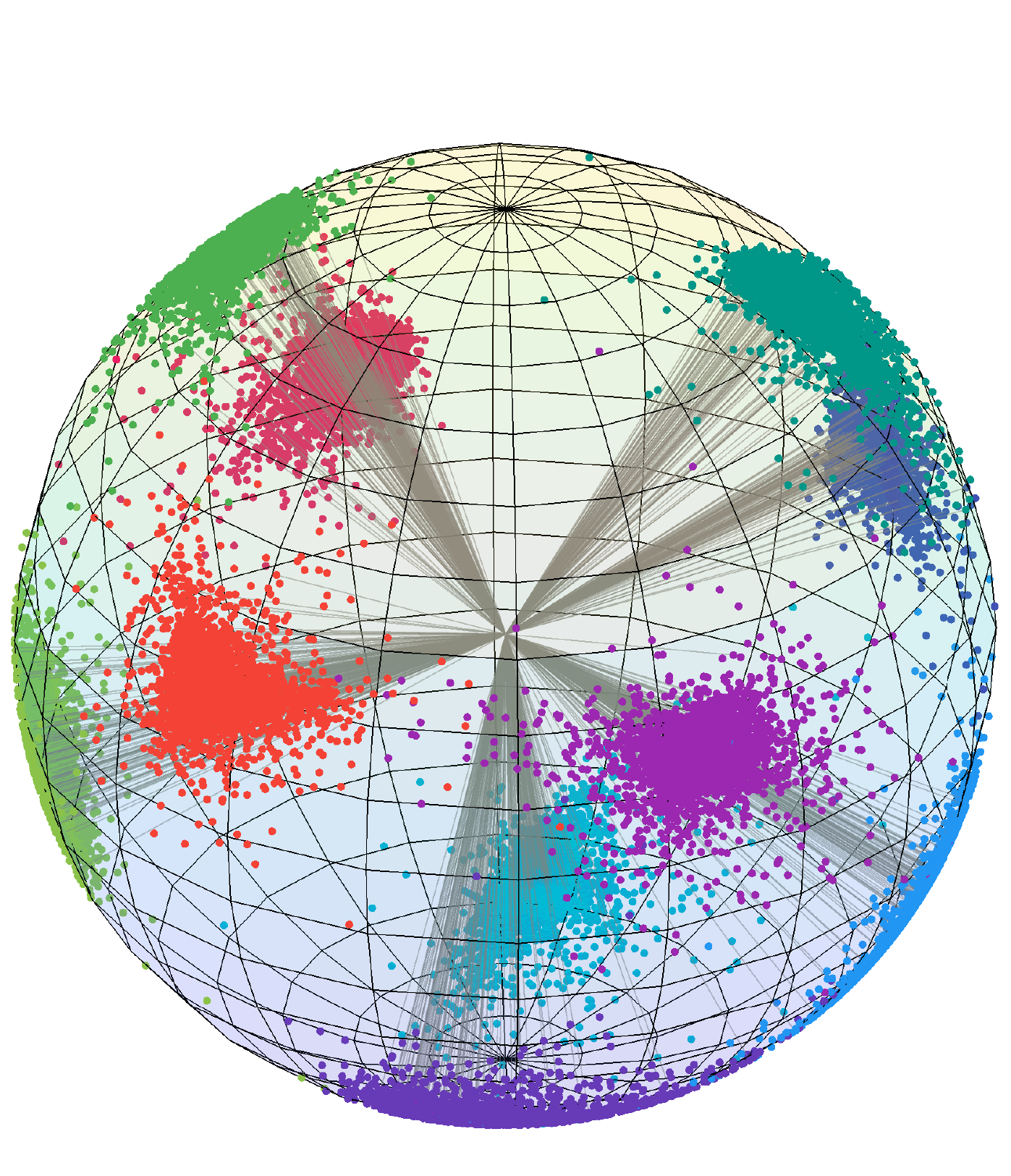}}
  %\vskip -0.1in
 \centerline{\hspace{-0.2in}(b) Training with $\mathcal{L}_{ce}\small{+}\mathcal{L}_{con}$}
  \end{minipage}
 % \vskip -0.05in
 \vspace{-10pt}
  \end{center} 
  \caption{Visualizations of  features learned by ResNet-18 on the CIFAR10 validation set. 
 Compared to training with only cross-entropy $\mathcal{L}_{ce}$, the contrastive loss $\mathcal{L}_{con}$ helps to regularize the feature space and make it more discriminative. Best viewed in color.}
  \vspace{-5pt}
  \label{visualization} 
\end{wrapfigure}
Please see Appendix A for the proof.
This theorem shows that $\mathcal{L}_{con}$ explicitly regularizes representation learning.   
On one hand, minimizing $\mathcal{L}_{con}$ will minimize $\mathcal{H}(Z|Y)$, which  encourages learning a low-entropy cluster for each class (\ie high intra-class compactness). On the other hand, minimizing $\mathcal{L}_{con}$ will maximize $\mathcal{H}(Z)$ and tends to learn a high-entropy  feature space (\ie large inter-class separation degree). This provides an additional regularization effect on the feature space, 
which can be observed by the feature visualization  in Figure~\ref{visualization}. 
As for the two assumptions, $\ell_2$-normalized features can be satisfied by a non-linear projection in practice (cf.~Section~\ref{sec:generation}), while  contrastive fine-tuning also empirically performs well on class-imbalanced datasets (cf.~Table~\ref{exp_inaturalist}).
Note that this analysis is different from the analysis in unsupervised contrastive learning~\cite{wang2020understanding}, which is specific to the (unlabeled) instance  level  rather than the class level.

% \begin{figure}[t]
% %\vskip -0.2in
%  \begin{minipage}{0.49\linewidth}
%  \centerline{\hspace{-0.15in}\includegraphics[width=3.2cm,clip,trim={4.3cm 7.7cm 2.5cm 5.5cm}]{figures/cifar_ce_dpi300.pdf}}
%  %\vskip -0.25in
%   \centerline{\hspace{-0.15in}(a) Training with $\mathcal{L}_{ce}$}
%  \end{minipage}
%  \hfill 
%  \begin{minipage}{0.49\linewidth}
%  \centerline{\hspace{-0.2in}\includegraphics[width=3.2cm,clip,trim={4.3cm 7.7cm 2.5cm 5.5cm}]{figures/cifar_contrastive_dpi300.pdf}}
%   %\vskip -0.1in
%  \centerline{\hspace{-0.2in}(b) Training with $\mathcal{L}_{ce}\small{+}\mathcal{L}_{con}$}
%   \end{minipage}
%  % \vskip -0.05in
%  \caption{Visualizations of learned features by ResNet-18 on the CIFAR10 valdiation set. 
%  Compared to training with only cross-entropy $\mathcal{L}_{ce}$, the contrastive loss $\mathcal{L}_{con}$ helps regularize the feature space and makes it more discriminative (\ie high intra-class compactness and large inter-class separation degree). Best viewed in color.} %\vskip -0.2in
%  \label{visualization} 
% \end{figure} 

\subsection{Optimization Effect of Contrastive Loss}\label{Sec:regularization2} 
We next show that the contrastive loss improves the optimization effectiveness during model training via Theorem~\ref{thm1}.

\begin{thm} \label{thm1}
Assuming the features are $\ell_2$-normalized and  the classes are balanced, the contrastive loss  is positive proportional  to  the infimum of conditional cross-entropy $\mathcal{H}(Y;\hat{Y}|Z)$, where the infimum is taken over classifiers: 
\begin{align}
    \mathcal{L}_{con}~\propto~ \inf  \underbrace{\mathcal{H}(Y;\hat{Y}|Z)}_{\text{Conditional CE}}  ~ - ~   \mathcal{H}(Y) \nonumber
\end{align}
\end{thm} 
Please see   Appendix A for proofs.
This theorem shows $\mathcal{L}_{con}$ boosts model optimization. % regarding cross-entropy. 
Concretely, the label $Y$ is given by datasets, so its entropy $\mathcal{H}(Y)$ is a constant and can be ignored. Hence, minimizing $\mathcal{L}_{con}$ will minimize the infimum of conditional cross-entropy $\mathcal{H}(Y;\hat{Y}|Z)$, which   provides  an additional optimization effect as compared to fine-tuning with only cross-entropy.  More intuitively, pulling positive pairs together and pushing negative pairs further apart make the predicted label distribution closer to the ground-truth distribution, which further minimizes the cross-entropy loss. 
 
%\blue{We conclude that integrating the contrastive loss into cross-entropy will offer additional benefits to fine-tuning in terms of both representation learning and model optimization.}

\section{Contrast-Regularized Tuning}

Based on the above theoretical analysis, we are motivated to  introduce  contrastive learning to fine-tune contrastive self-supervised visual models on downstream tasks. Nevertheless, we empirically find that simply adding the contrastive loss to the fine-tuning objective is insufficient to obtain promising performance (cf.~Table~\ref{ablation}). One key cause is that contrastive
learning highly relies on positive/negative sample pairs, but the majority of samples are easy-to-contrast pairs~\cite{harwood2017smart,wu2017sampling} that  may produce negligible contrastive loss gradients. This makes contrastive learning fail to learn more  discriminative representations and thus suffer from unsatisfactory performance. To address this issue and better fine-tune contrastive self-supervised models, we propose a new contrast-regularized tuning (Core-tuning) method based on a novel  hard sample pair mining strategy as follows.

%.  First,  hard sample pairs are informative for contrastive learning but the majority of training samples are easy to contrast~\cite{wu2017sampling}, so it is important to  mine hard pairs for more effective fine-tuning. 
  %Since the hard pairs are more informative for contrastive learning~\cite{harwood2017smart},  
%To handle these challenges, Core-tuning applies two novel designs. First, Core-tuning generates both hard positive   and  hard negative pairs  for more effective contrastive fine-tuning via a simple yet effective feature mixup strategy (see Section~\ref{sec:generation}). 
 %(see Section~\ref{sec:weight}). 

\subsection{Hard Sample Pair Mining for Contrastive Fine-Tuning}\label{sec:generation} 
For more effective contrastive fine-tuning, Core-tuning generates both hard positive   and  hard negative pairs  via a new  hardness-directed mixup strategy, and meanwhile assigns higher importance  weights  to hard positive pairs  via a new focal contrastive loss.

\textbf{Hard positive pair generation.}
As shown in Figure~\ref{challenges_motivations} (b), for a given feature anchor $z_i$,  we first find its hardest positive data $(z_i^{hp},y_i^{hp})$ and hardest negative data  $(z_i^{hn},y_i^{hn})$ based on cosine similarity. 
That is, $z_i^{hp}$ is the positive data (from the same class) with the lowest cosine similarity to the anchor, and $z_i^{hn}$ is the negative data (from different classes) most similar to the anchor.
We then generate a hard positive pair as a convex combination of the two hardest pairs:
\begin{align} 
 z_{i}^+=\lambda z_i^{hp} + (1-\lambda) z_i^{hn}; ~~~~
 y_{i}^+=  \lambda   y_i^{hp} +   (1-\lambda) y_i^{hn}, \nonumber
\end{align} 
where $\lambda \small{\sim} \text{Beta}(\alpha,\alpha)\small{\in} [0,1]$~\cite{zhang2018mixup}, in which $\alpha \small{\in} (0, \infty)$ is a hyper-parameter to decide the Beta distribution. 
The generated positive pairs are located between positives and negatives and thus are harder to contrast. Note that the generated positive pairs do not have to be   the hardest. In fact, as long as we can generate relatively hard pairs, the performance of contrastive fine-tuning could be improved.
%We denote the generated hard positive set by $\mathcal{B}^+ \small{=}\{z_{i}^+\}_{i=1}^n$.   $\text{Beta}(\alpha,\alpha)$
 
\textbf{Hard negative pair generation.}
As shown in Figure~\ref{challenges_motivations} (c), for a given feature anchor $z_i$, we randomly select a negative sample $(z_i^n,y_i^n)$ to synthesize a semi-hard negative pair as follows: 
\begin{align} 
 z_{i}^-= (1-\lambda)  z_i + \lambda  z_i^n; ~~~~
 y_{i}^-= (1-\lambda)  y_i + \lambda y_i^n, \nonumber
\end{align}
where $\lambda \sim \text{Beta}(\alpha,\alpha)$.
The reason why we  select a random negative sample instead of  the hardest negative  is that generating too hard negatives may result in false negatives and degrade performance.
Note that semi-hard negatives may  even yield better performance in metric learning~\cite{wu2021conditional}.
%We denote the generated hard negative  set by $\mathcal{B}^- \small{=}\{(z_{i}^-,y_i^-)\}_{i=1}^n$.  

\textbf{Hard pair reweighting.}\label{sec:weight}
After generating hard sample pairs, we use  an additional two-layer MLP head $G_c$ to obtain $\ell_2$-normalized contrastive features $v_i \small{=} G_c(z_i)/\|G_c(z_i)\|_2$, since a nonlinear projection improves contrastive learning~\cite{chen2020big,chen2020improved}.
Based on these features, one may directly use $\mathcal{L}_{con}$ in Eq.~(\ref{eq1}) for fine-tuning. 
However, since hard pairs are more informative for contrastive learning, we propose to assign higher importance weights to hard positive pairs.
Inspired by focal loss~\cite{lin2017focal}, we find  that hard positive pairs generally lead to a low prediction probability $p_{ij} \small{=}\frac{\exp (v_i^{\top} v_j/\tau)}{\sum_{v_k\in A_i}\exp (v_i^{\top} v_k/\tau)}$.
Thus, we reweight 
$\mathcal{L}_{con}$ with $(1\small{-}p_{ij})$ and develop a focal contrastive loss:
\begin{align} 
    \mathcal{L}_{con}^f \small{=}\small{-}\frac{1}{n}\sum_{i=1}^n\frac{1}{|P_i|}\sum_{v_j\in P_i} (1\small{-}p_{ij}) \log \frac{e^{(v_i^{\top} v_j/\tau)}}{\sum_{v_k\small{\in} A_i}e^{(v_i^{\top} v_k/\tau)}}, \nonumber
\end{align}
where $P_i$, $A_i$ denote the anchor's positive and full pair sets,   which contain the generated hard pairs. Via the  hard pair mining  strategy, Core-tuning is able to learn a more  discriminative feature space.

%Second, Core-tuning further uses the mixed features for classifier training \wrt cross-entropy $\mathcal{L}_{ce}^m$, so that the learned  decision boundary is smoother (see Section~\ref{sec:smmoth}).

\subsection{Overall Training Scheme and Smooth Classifier Learning}\label{sec:smmoth} 

In fine-tuning, both the feature extractor and classifier need to be trained, so  the final training scheme of Core-tuning\footnote{The pseudo code is provided in the supplementary.}  is to minimize the following objective:
\begin{align} 
\min ~~~ \underbrace{\mathcal{L}_{ce}^m}_{\text{cross-entropy loss}}+  \underbrace{\eta\mathcal{L}_{con}^{f}}_{\text{focal contrastive loss}}, \nonumber
\end{align} 
where $\eta$ is a trade-off factor. Since hard sample mining has helped to learn a  discriminative feature space, the remaining question is how to train the classifier for better exploiting such a feature space.

\textbf{Smooth classifier learning.} Previous work~\cite{verma2019manifold} has found  that the classifier simply trained with cross-entropy is often sharp and close to data. This may make the classifier   fail to  exploit the high  inter-class separation degree in the discriminative feature space  due to closeness to training data, as well as suffer from limited generalization performance since the classifier near the training data may lead to incorrect yet confident predictions when evaluated on slightly different test samples.
To address this, inspired by the effectiveness of mixup for helping learn a smoother decision boundary~\cite{mangla2020charting,verma2019manifold}, we further use the mixed data from the generated hard sample pair set (denoted by $\mathcal{B}$) for classifier training:   
$\mathcal{L}_{ce}^m \small{=}\small{-} \frac{1}{n}\sum_{i=1}^n   {y_{i}} 
  \log(\hat{y}_i)\small{-}\frac{1}{|\mathcal{B}|}\sum_{(z_j,y_j)\in \mathcal{B}}   y_j \log(G_y(z_j))$.  
In this way, Core-tuning is able to learn a smoother classifier that is far away from the training data, and thus can better exploit the learned discriminative feature space and improve the model generalizability.

%\begin{align}%\label{eq:classification}
%  \mathcal{L}_{ce}^m \small{=}\small{-} \frac{1}{n}\sum_{i=1}^n   {y_{i}} 
%   \log(\hat{y}_i)\small{-}\frac{1}{|\mathcal{B}|}\sum_{(z_j,y_j)\in \mathcal{B}}   y_j \log(G_y(z_j)). \nonumber
% \end{align}

% One may also consider using the hard positive set $\mathcal{B}^+$.
% Nevertheless, the mixed positives are located in the borderline area between positives and negatives, which has already been  covered by the mixed negatives.
% Further using $\mathcal{B}^+$ will not bring extra performance improvement. 

\section{Experiments}  
We first test the effectiveness of Core-tuning  on image classification and then apply it to semantic segmentation. Next, since Core-tuning potentially improves model generalizability, we further study how it affects model generalization to new domains and model robustness to adversarial samples.

\begin{table*}[t] 
%\vspace{-0.05in}
	\caption{Comparisons of various fine-tuning methods for the MoCo-v2 pre-trained ResNet-50 model  on image classification in terms of top-1 accuracy. SL-CE-tuning denotes supervised pre-training on ImageNet and then fine-tuning with cross-entropy.} 
     \label{exp_classification} 
    \begin{center}
     \scalebox{0.73}{  
    \begin{threeparttable} 
	\begin{tabular}{lcccccccccc}\toprule
        Method  &   ImageNet20 & CIFAR10 & CIFAR100 & Caltech101 & DTD  &Aircraft& Cars & Pets & Flowers  & Avg. \\ \toprule   
        SL-CE-tuning & 91.01 &94.23 & 83.40 & 93.39 & 74.40  &87.03 &  89.77 &  92.17 & 98.78 & 89.35 \\     \midrule     
        CE-tuning   & 88.28  & 94.70 &80.27  & 91.87  &   71.68 &  86.87  &88.61  & 89.05  & 98.49   & 87.76 \\  
        L2SP~\cite{li2018explicit}    &  88.49 & 95.14  &81.43  &  91.98 & 72.18  & 86.55 & 89.00 &  89.43 &  98.66  &  88.10\\    
        M$\&$M~\cite{zhan2018mix}    & 88.53 & 95.02 & 80.58  & 92.91 & 72.43  & 87.45 & 88.90  &  89.60 & 98.57   & 88.22 \\  
        DELTA~\cite{li2019delta}  & 88.35  & 94.76 &    80.39  & 92.19   & 72.23   &87.05  & 88.73 & 89.54 &  98.65  & 87.99  \\
         BSS~\cite{chen2019catastrophic}  & 88.34 & 94.84 & 80.40 &  91.95 & 72.22    & 87.18  & 88.50    & 89.50  & 98.57 & 87.94\\  
        RIFLE~\cite{li2020rifle}     &89.06 & 94.71& 80.36 &  91.94 & 72.45  & 87.60  &89.72   & 90.05 &98.70  & 88.29\\  
         SCL~\cite{gunel2020supervised}   &   89.29 & 95.33  & 81.49  &  92.84 & 72.73  & 87.44  & 89.37 &  89.71 &  98.65  & 88.54\\  
        Bi-tuning~\cite{zhong2020bi}   & 89.06 & 95.12 & 81.42 &  92.83 & 73.53  &87.39  &89.41   & 89.90  & 98.57  & 88.58\\          
        Core-tuning (ours)& \textbf{92.73}& \textbf{97.31} & \textbf{84.13} &  \textbf{93.46} & \textbf{75.37}   & \textbf{89.48}  & \textbf{90.17}   & \textbf{92.36}  & \textbf{99.18} & \textbf{90.47} \\ 
 
        \bottomrule
	\end{tabular} 
    \end{threeparttable}}
    \end{center}  
   \vspace{-0.1in}
\end{table*}

\subsection{Results on Image Classification}  
\textbf{Settings.} 
As there is no fine-tuning method devoted to contrastive self-supervised models, we compare Core-tuning with advanced fine-tuning methods for general models (\eg supervised pre-trained models): L2SP~\cite{li2018explicit}, M$\&$M~\cite{zhan2018mix}, DELTA~\cite{li2019delta}, BSS~\cite{chen2019catastrophic}, RIFLE~\cite{li2020rifle}, SCL~\cite{gunel2020supervised} and Bi-tuning~\cite{zhong2020bi}. We denote  the fine-tuning with cross-entropy by CE-tuning.
 
Following~\cite{kornblith2019better}, we test on 9 natural image datasets, including ImageNet20 (a subset of ImageNet with 20 classes), CIFAR10, CIFAR100~\cite{krizhevsky2009learning}, Caltech-101~\cite{fei2004learning}, DTD~\cite{cimpoi2014describing}, FGVC Aircraft~\cite{maji2013fine}, Standard Cars~\cite{krausecollecting}, Oxford-IIIT Pets~\cite{parkhi2012cats} and Oxford 102 Flowers~\cite{nilsback2008automated}.  Specifically, ImageNet20 is an ImageNet subset with 20 classes, by combining the ImageNette and ImageWoof datasets~\cite{imagewang}. Here, we do not directly test on ImageNet~\cite{deng2009imagenet}, since all CSL models are pre-trained on the ImageNet dataset. These datasets cover a wide range of fine/coarse-grained object recognition tasks.

We implement Core-tuning in PyTorch\footnote{The source code of Core-tuning is available at: \url{https://github.com/Vanint/Core-tuning}.}. Following~\cite{ericsson2021well}, we use ResNet-50 (1$\times$), pre-trained  by various CSL methods on ImageNet, as the network backbone. All checkpoints of pre-trained models are  provided by authors or by the PyContrast repository\footnote{\url{https://github.com/HobbitLong/PyContrast}.}. 
Following~\cite{chen2020simple}, we perform parameter tuning for $\eta$ and $\alpha$ from $\{0.1, 1, 10\}$ on each dataset. Moreover, we set the temperature $\tau\small{=}0.07$.
To make the generated negative pairs closer to negatives, we clip $\lambda\small{\sim}\text{Beta}(\alpha,\alpha)$ by $\lambda \small{\geq}\lambda_n$ when generating hard negative pairs, where $\lambda_{n}$ is a threshold and we set it to 0.8. 
% To save computation costs, we only use the mixed hard negative samples for classification training.
All results are averaged over 3 runs in terms of the top-1 accuracy.
More dataset details, more implementation details and the parameter analysis are put in  Appendices C and E. 
%\footnote{https://github.com/google-research/simclr.}  

\textbf{Comparisons with previous methods.} 
We report the fine-tuning performance of the MoCo-v2 pre-trained model in Table~\ref{exp_classification}. 
When using the standard CE-tuning,  the MoCo-v2 pre-trained model performs worse than the supervised pre-trained model on most datasets.
This is because the self-supervised pre-trained model is less class-discriminative than the supervised pre-trained model due to the lack of annotations during pre-training. 
Moreover, the classic fine-tuning methods designed for supervised pre-trained models (\eg L2SP and DELTA)  cannot fine-tune the contrastive self-supervised model very well.
One reason is that the contrastive pre-training  task is essentially different from the downstream classification task, so strictly regularizing the difference between the contrastive self-supervised model and the fine-tuned model may  lead  to negative/poor transfer. 
In addition, M$\&$M, SCL and Bi-tuning use the triplet loss  or the  contrastive loss during fine-tuning. 
However, they ignore the two challenges in contrastive fine-tuning as mentioned in Figure~\ref{challenges_motivations}, leading to limited model performance on downstream tasks.  
In contrast, Core-tuning handles those challenges well and improves the fine-tuning performance of CSL models a lot.
This result demonstrates the superiority of Core-tuning. More results like the standard error are put in Appendix D. 
 
\begin{table*}[t]  
	\vspace{-0.1in}
	\centering 
	\caption{Ablation studies of Core-tuning (Row 5) for fine-tuning MoCo-v2 pre-trained ResNet-50 in terms of top-1 accuracy, where cross-entropy is used in all baselines.  Here, $\mathcal{L}_{con}$ is the original contrastive loss, while $\mathcal{L}^f_{con}$ is our focal contrastive loss. Moreover,  ``mix" denotes the manifold mixup, while ``mix-H" indicates the proposed hardness-directed mixup strategy in our method.} %\vspace{-0.1in}
     \label{ablation} 
    \begin{center}
     \scalebox{0.68}{  
    \begin{threeparttable} 
	\begin{tabular}{cc|cc|cccccccccc}\hline
         $\mathcal{L}_{con}$  &$\mathcal{L}^f_{con}$ & mix   & mix-H &   ImageNet20 &CIFAR10 & CIFAR100 & Caltech101 & DTD  &Aircraft& Cars & Pets & Flowers  & Avg. \\\hline    
        %SL-CE & 94.23 & 83.40 & 93.65 & 74.40  &87.03 &  89.77 &  92.17 & 98.78  \\       
         &  & &   & 88.28  & 94.70 &80.27  & 91.87  &   71.68 &  86.87  &88.61  & 89.05  & 98.49   & 87.76 \\  
               $\surd$  &  & & &  89.29 & 95.33  & 81.49  &  92.84 & 72.73  & 87.44  & 89.37 &  89.71 &  98.65  & 88.54\\ 
         &  & $\surd$ &  &90.67  & 95.43 &    81.03  & 92.68   & 73.31   &88.37  & 89.06 & 91.37 &  98.74  & 88.96  \\

       $\surd$  &   & & $\surd$ & 92.20 & 97.01 & 83.89 &  93.22 & 74.78    &88.88  &89.79    & 91.95  & 98.94  & 90.07\\  
          &  $\surd$  &  & $\surd$ & \textbf{92.73}& \textbf{97.31} & \textbf{84.13} &  \textbf{93.46} & \textbf{75.37}   & \textbf{89.48}  & \textbf{90.17}   & \textbf{92.36}  & \textbf{99.18} & \textbf{90.47} \\  
        \hline
	\end{tabular} 
    \end{threeparttable}}
    \end{center}  \vspace{-0.1in} 
\end{table*}
 
\textbf{Ablation studies of Core-tuning.}
We conduct ablation studies for Core-tuning regarding the focal contrastive loss and  the   hardness-directed mixup strategy. As shown in Table 2, each component improves the fine-tuning performance in Core-tuning. 
Note that the mixup in Row 3 is the manifold mixup~\cite{verma2019manifold}, which is essentially designed for classification and is expected to outperform our hardness-directed mixup strategy regarding classification performance. However, our proposed Core-tuning (Row 5) still shows obvious improvement on all datasets, which strongly verifies the value of contrastive fine-tuning. More ablation results for verifying the effectiveness of hard pair generation and  smooth classifier learning  are put in Appendix E.

\textbf{Results on different pre-training methods.}
In previous experiments, we fine-tune the MoCo-v2 pre-trained ResNet-50, but it is unclear whether Core-tuning can be applied to fine-tune   models with other pre-training methods. Hence, we further use Core-tuning to fine-tune  ResNet-50, pre-trained by other CSL methods (\ie InsDis~\cite{wu2018unsupervised}, PIRL~\cite{misra2020self}, MoCo-v1~\cite{he2020momentum} and InfoMin~\cite{tian2020makes}), 
non-contrastive self-supervised methods (\ie DeepCluster-v2~\cite{caron2018deep}, SwAV~\cite{caron2020unsupervised} and BYOL~\cite{grill2020bootstrap}), and supervised learning. As shown in Table~\ref{exp_ssl_model}, Core-tuning  fine-tunes all   pre-trained models consistently better than  CE-tuning on 3 image classification datasets. Such results verify the generalizability of the proposed Core-tuning. More results on different pre-trained models are put in Appendix D.

	\begin{table}[t!]
	\begin{minipage}{0.55\textwidth}
		\caption{Fine-tuning results of ResNet-50, pre-trained by various methods. ``Cont." indicates contrastive self-supervised pre-training; CE indicates cross-entropy.}
     \label{exp_ssl_model} 	\vspace{-0.1in}
    \begin{center}
     \scalebox{0.6}{  
    \begin{threeparttable} 
	\begin{tabular}{lccccccccc}\toprule
	 \multirow{2}{*}{Pre-training}   &   \multirow{2}{*}{Types}   &   \multicolumn{2}{c}{Caltech101} &&  \multicolumn{2}{c}{DTD} &&  \multicolumn{2}{c}{Pets} \cr  \cmidrule{3-4}    \cmidrule{6-7} \cmidrule{9-10}  
     &  & CE   & ours && CE   & ours  && CE   & ours \cr
     \toprule
      InsDis~\cite{wu2018unsupervised} &  \multirow{4}{*}{Cont.} &  82.30 & \textbf{88.60}    &&69.81 & \textbf{70.94}  && 87.57& \textbf{89.59} \\    
       PIRL~\cite{misra2020self}&   &  84.23 &\textbf{89.29}    && 68.95 & \textbf{71.72} && 86.87 & \textbf{89.52} \\    
       MoCo-v1~\cite{he2020momentum} & &85.74 & \textbf{89.16}   &&69.91 &\textbf{71.90} && 88.16 & \textbf{90.11} \\ 
      % MoCo-v2~\cite{chen2020improved}   & &91.87 & \textbf{93.46}   &&71.68 & \textbf{75.37}  && 89.05 & \textbf{92.36} \\
        %SimCLR-v1 & 90.53 &\textbf{92.40}  &&69.93 & \textbf{71.26}  && 89.34 & \textbf{90.89} \\ 
     %&   SimCLR-v2 &92.44 & \textbf{93.37}   && 71.63 & \textbf{74.75}  && 88.28 & \textbf{90.64}\\ 
      InfoMin~\cite{tian2020makes} & &  92.73 & \textbf{94.01}   && 72.59 & \textbf{74.89} && 90.00 & \textbf{92.34} \\   \midrule
       DeepCluster\cite{caron2018deep} & \multirow{3}{*}{Non-Cont.}  &  89.99 & \textbf{92.34}   && 72.77 & \textbf{75.21} && 90.53 & \textbf{93.17} \\ 
       SwAV~\cite{caron2020unsupervised}  & &  87.71 & \textbf{91.34}   && 75.29 & \textbf{77.41} && 92.48 & \textbf{93.29} \\  
        BYOL~\cite{grill2020bootstrap}  & &  91.19 & \textbf{93.25}   && 74.94 & \textbf{76.56} && 92.39 & \textbf{93.74} \\  
       
          \midrule
        CE  & Supervised &  93.65 & \textbf{94.20}   && 74.40 & \textbf{77.27} && 92.17 & \textbf{93.82} \\ 
        \bottomrule
	\end{tabular} 
   \end{threeparttable}}
    \end{center}  %\vskip -0.1in 
	\end{minipage}
	~~
	\begin{minipage}{0.42\textwidth}
	\caption{Fine-tuning performance of various architectures. Here,    ResNet (R) and  ResNeXt (RX) are pre-trained by InfoMin; DeiT-S~\cite{touvron2021training} is pre-trained by DINO~\cite{caron2021emerging}. }%CE indicates cross-entropy. 
    \vspace{-0.15in}
     \label{exp_architecture} 
    \begin{center}
     \scalebox{0.6}{  
    \begin{threeparttable} 
	\begin{tabular}{lcccccccc}\toprule
	  \multirow{2}{*}{Archs.}   &   \multicolumn{2}{c}{Caltech101} &&  \multicolumn{2}{c}{DTD} &&  \multicolumn{2}{c}{Pets} \cr  \cmidrule{2-3}    \cmidrule{5-6} \cmidrule{8-9}  
       & CE   & ours && CE   & ours  && CE   & ours \cr
     \toprule
        R-50 & 92.73 & \textbf{94.01}   && 72.59 & \textbf{74.89} && 90.00 & \textbf{92.34} \\   
        R-101 &  93.06 &\textbf{94.33}    && 73.38 & \textbf{75.09} && 90.84 & \textbf{92.91} \\    
        %MoCo-v1  &85.74 & \textbf{89.16}   &&69.91 &\textbf{71.90} && 88.16 & \textbf{90.11} \\ 
        R-152 &93.39 & \textbf{94.66}   &&73.74 & \textbf{75.42}  && 91.08 & \textbf{92.97} \\
        %SimCLR-v1 & 90.53 &\textbf{92.40}  &&69.93 & \textbf{71.26}  && 89.34 & \textbf{90.89} \\ 
        RX-101 &93.71 & \textbf{95.12}   && 74.43 & \textbf{75.97}  && 91.97& \textbf{94.04}\\ 
        RX-152 & 93.92 & \textbf{95.19}   && 74.76 & \textbf{76.22} && 92.70 & \textbf{94.49} \\ 
        \midrule
        %MNet-v2 & 88.96 & \textbf{92.07}   && 68.07 & \textbf{71.77} && 87.98 & \textbf{92.77} \\ 
        DeiT-S/16 & 91.24 & \textbf{92.31}   &&71.35 & \textbf{72.83} &&  92.43 & \textbf{93.72} \\ 
        \bottomrule
	\end{tabular} 
   \end{threeparttable}}
    \end{center}  %\vskip -0.2in  
    	\end{minipage}
    \vspace{-0.1in}
    \end{table}
	%MNet indicates the supervised pre-trained MobileNet~\cite{sandler2018mobilenetv2}, while 

\textbf{Results on different network architectures.} 
Previous experiments are based on ResNet-50, while it is unclear whether Core-tuning can be applied to other network architectures. Hence, we further use Core-tuning to fine-tune various residual network  architectures (\ie ResNet-101 and 152; ResNeXt-101 and 152~\cite{xie2017aggregated}) pre-trained by InfoMin~\cite{tian2020makes}, and vision transformer (\ie DeiT-S/16~\cite{touvron2021training}) pre-trained by DINO~\cite{caron2021emerging}. As shown in Table~\ref{exp_architecture}, Core-tuning fine-tunes all network architectures well on all three datasets, showing strong universality.

\textbf{Results on different data sizes.}
The labeled data may be scarce in downstream tasks. Hence, we further evaluate Core-tuning on ImageNet20 with  different sampling rates  of data. We report the results in Table~\ref{exp_diverse_number},  while the results on the full ImageNet20   have been listed in Table~\ref{exp_classification}. Specifically, Core-tuning outperforms  baselines in all cases.
Note that when the data is very scarce (\eg 10$\%$), the fine-tuning performance of CE-tuning degrades and fluctuates  significantly, in which case  Core-tuning obtains more significant  improvement and achieves more stable performance. 

\textbf{Results on large-scale and class-imbalanced dataset.} The real-world datasets may be large-scale and class-imbalanced~\cite{zhang2021test,zhang2021deep,zhao2018adaptive}, so we also evaluate Core-tuning on a long-tailed iNaturalist18 dataset~\cite{van2018inaturalist}, consisting 437,513 images from 8,142 classes. 
As shown in Table~\ref{exp_inaturalist}, Core-tuning also performs well on the large-scale and class-imbalanced dataset for fine-tuning contrastive pre-trained models. Note that in our theoretical analysis, we assume that the classes are balanced with  the same data number to facilitate analysis. Nevertheless, this assumption does not mean that contrastive fine-tuning cannot handle class-imbalanced datasets. Here, the promising results on iNaturalist-18 verify the effectiveness of Core-tuning on highly class-imbalanced scenarios.

%We adopt the official training and validation splits for the experiments. 
%\blue{The labeled data may be scarce in downstream tasks. Hence, we further evaluate Core-tuning on ImageNet20 with  different sampling rates  of data. We report the results in Table~\ref{exp_inaturalist}. Specifically, Core-tuning outperforms  baselines in all cases.}
 
	\begin{table}[t!]
	\vspace{-0.1in}
	\begin{minipage}{0.55\textwidth}
	\caption{Fine-tuning  performance of the MoCo-v2 pre-trained ResNet-50 with various numbers of labeled data.}
     \label{exp_diverse_number} \vspace{-0.05in}
    \begin{center}
     \scalebox{0.685}{  
    \begin{threeparttable} 
	\begin{tabular}{lcccc}\toprule
	   \multirow{2}{*}{Method}   &    \multicolumn{4}{c}{Sampling Rates on ImageNet20}  \cr  \cmidrule{2-5}    
        & $10\%$ & $25\%$ & $50\%$ & $75\%$   \\\toprule     
        %SL-CE & 94.23+/-0.07 & 83.40+/-0.12& 93.65+/-0.21 & 74.40+/-0.45 \\    
        CE-tuning &  52.97+/-3.96 & 63.17+/-3.94 & 81.78+/-1.37 & 85.85+/-0.11  \\     
        Bi-tuning  &60.50+/-1.11  &   75.86+/-0.74  & 83.18+/-0.27  & 87.19+/-0.19   \\
        Core-tuning & \textbf{78.64+/-0.58}  & \textbf{84.48+/-0.34}  &  \textbf{89.09+/-0.40} & \textbf{90.93+/-0.24}  \\   
        \bottomrule
	\end{tabular}
	\end{threeparttable}}
    \end{center} %  \vskip -0.1in
    
     \vspace{0.07in}
	\caption{Fine-tuning  performance of the MoCo-v2 pre-trained ResNet-50 on   large-scale and class-imbalanced iNaturalist18   in terms of top-1 accuracy. } 
     \label{exp_inaturalist} 
    \begin{center} \vspace{-0.05in}
     \scalebox{0.87}{  
    \begin{threeparttable} 
	\begin{tabular}{lc}\toprule
        Fine-tuning method~~~~  &   iNaturalist18 \\ \toprule   
      %  SL-CE-tuning & 91.01 \\     
        CE-tuning   & 61.72+/-0.18  \\   
        CE-Contrastive-tuning   & 62.75+/-0.22 \\          
        Core-tuning (ours)~~~~~~& \textbf{63.57+/-0.09} \\ 
 
        \bottomrule
	\end{tabular} 
    \end{threeparttable}}
    \end{center}  
	\end{minipage}
	~~
	\begin{minipage}{0.42\textwidth}
	\caption{Fine-tuning performance on PASCAL VOC semantic segmentation based on DeepLab-V3 with ResNet-50, pre-trained by various CSL methods.  }  %CE indicates cross-entropy.
     \label{exp_segmentation}  \vspace{-0.17in}
    \begin{center}
    \scalebox{0.72}{  
    \begin{threeparttable} 
	\begin{tabular}{lcccc}\toprule
        Pre-training  & Fine-tuning&  MPA & FWIoU  & MIoU \\ \toprule     
        Supervised  &  CE & 87.10 & 89.12 & 76.52  \\  
        \midrule    
        \multirow{2}{*}{InsDis}  & CE & 83.64 & 88.23  & 74.14 \\    
          & ours &  \textbf{84.53}  & \textbf{88.67} & \textbf{74.81}\\    
          %\hline
         \midrule    
        \multirow{2}{*}{PIRL}  & CE & 83.16 & 88.22  & 73.99\\    
          & ours & \textbf{85.30}  & \textbf{88.95} &\textbf{75.49}      \\   \midrule   
     \multirow{2}{*}{MoCo-v1}  & CE & 84.71 & 88.75 & 74.94\\    
          & ours & \textbf{85.70}  &\textbf{89.19}  & \textbf{75.94} \\    
          \midrule    
           
    \multirow{2}{*}{MoCo-v2}  & CE & 87.31 & 90.26 & 78.42\\    
          & ours & \textbf{88.76}  &\textbf{90.75}  & \textbf{79.62} \\    
          \midrule    
      
        % \multirow{2}{*}{SimCLR-v2}  & CE  & 87.37 & 90.27  & 78.16 \\    
        %   & ours & \textbf{87.95}  & \textbf{90.71}  & \textbf{79.15} \\    \midrule   
      
      \multirow{2}{*}{InfoMin}  & CE & 87.17 & 89.84  & 77.84 \\    
          & ours & \textbf{88.92}  &\textbf{90.65}  & \textbf{79.48}      \\  
 
        \bottomrule
	\end{tabular}
	   \end{threeparttable}}
    \end{center}   
    	\end{minipage}
    	\vspace{-0.05in}
    \end{table}

% \newpage
\subsection{Results on Semantic Segmentation}
We next apply Core-tuning to fine-tune contrastive self-supervised models on semantic segmentation.
  
\textbf{Implementation details.}
We adopt the DeepLab-V3 framework~\cite{chen2017rethinking} for PASCAL VOC semantic segmentation and use   CSL pre-trained ResNet-50 models as the backbone. In Core-tuning, we enforce the contrastive regularizer after the penultimate layer of ResNet-50 via an additional global average pooling.  
%Due to the characteristics of dense prediction,  we only generate hard positive pairs and do not use the mixed data for classifier training.
Following~\cite{wang2021dense}, the model is fine-tuned on VOC train$\_$aug2012 set for 30k steps via SGD  based on two  GPUs and evaluated on val2012 set. The image is rescaled to $513 \small{\times} 513$ with random crop and flips for training  and with center crop for evaluation. The batch size  and   output stride are 16. Besides, we set the initial learning rate to 0.1 and adjust it via the poly decay schedule. 
%The momentum parameter is set to 0.9, and the   factor of weight decay is set to $10^{-4}$. 
Other parameters are the same as image classification.  We use three metrics: Mean Pixel Accuracy~(MPA), Frequency Weighted Intersection over Union~(FWIoU) and Mean  Intersection over Union~(MIoU).

\textbf{Results.}
As shown in Table~\ref{exp_segmentation},  Core-tuning  contributes to the fine-tuning performance of all CSL models  in terms of MPA, FWIoU and MIoU. The promising results demonstrate  the effectiveness of Core-tuning on semantic segmentation.  
Interestingly, we find that with standard fine-tuning, the models pre-trained by MoCo-v2 and InfoMin  have already outperformed the supervised pre-trained model. One explanation is that self-supervised pre-training may keep more visual information, compared to supervised pre-training that mainly extracts information specific to classification~\cite{zhao2020makes}. In other words, unsupervised contrastive learning may extract more beneficial information  for dense prediction, which inspires us to explore unsupervised contrastive regularizers in the future. %The results with the standard error are put in Appendix D. 

\subsection{Effectiveness on Cross-Domain Generalization}
The generalizability of deep networks to unseen domains is important for their application to real-world scenarios~\cite{dou2019domain}. We thus wonder  whether Core-tuning also benefits model generalization    on downstream tasks, so we apply Core-tuning to the task of domain generalization (DG).

\textbf{Implementation details.}   
DG aims to train a model on multiple source domains and expect it to generalize well to an unseen target domain. Specifically, we use MoCo-v2 pre-trained ResNet-50 as the backbone, and evaluate  Core-tuning on 3 benchmark datasets, \ie PACS~\cite{li2017deeper}, VLCS~\cite{fang2013unbiased} and Office-Home~\cite{venkateswara2017deep}. For training, we use Adam optimizer with batch size 32. The learning rate is set to $5\small{\times}10^{-5}$ and the training step is 20,000. More implementation details are put in Appendix C. 
   
\textbf{Results.} 
We report the results on PACS and  VLCS in Table~\ref{exp_generalization} and the results on OfficeHome in Appendix D, from which we draw observations as follows. First,  when fine-tuning with cross-entropy, the contrastive self-supervised model  performs  worse than the supervised pre-trained model. This results from the relatively worse discriminative abilities of the contrastive self-supervised model, which can also be found in  Table~\ref{exp_classification}. Second, enforcing the contrastive regularizer during fine-tuning improves DG performance, since the contrastive  regularizer helps to learn more discriminative features (cf.~Theorem~\ref{thm0}) and also helps to alleviate distribution shifts among  domains~\cite{kang2019contrastive}. Last, Core-tuning further improves the generalization performance of models. This is because hard pair generation further boosts contrastive learning,  while smooth classifier learning  benefits model generalizability. We thus conclude that Core-tuning is beneficial to model generalization on downstream tasks.

\begin{table}[t]   
\vspace{-0.1in}
\caption{Domain generalization accuracies of various fine-tuning methods for MoCo-v2 pre-trained ResNet-50. CE means cross-entropy;  CE-Con enhances CE with the contrastive loss. Moreover, A/C/P/S and C/L/V/S are different domains in PACS and VLCS datasets, respectively.}  \vspace{-0.05in}
    \label{exp_generalization} 
    \begin{center}
    \scalebox{0.7}{  
    \begin{threeparttable} 
	\begin{tabular}{ccccccccccccc}\toprule
        \multirow{2}{*}{Pre-training}  &  \multirow{2}{*}{Fine-tuning}  &    \multicolumn{5}{c}{PACS}  &&     \multicolumn{5}{c}{VLCS} \cr    \cmidrule{3-7} \cmidrule{9-13} 
        &  & A   & C  & P   &  S & Avg.  &&  C   & L  & V   &  S & Avg. \\\toprule    
        Supervised &  CE & 83.65  &  79.21 & 96.11  & 81.46  &  85.11  &&  98.41    & 63.81  & 68.55 & 75.45 & 76.56  \\   \midrule 
        \multirow{3}{*}{MoCo-v2} & CE & 78.71 & 76.92 & 90.87 &  75.67   & 80.54 &&   94.96  & 66.87   & 68.96 &  64.98 & 73.94  \\    
          & CE-Con & 85.11  &81.77 & 95.58  & 80.12  & 85.65  &&  95.94 &  67.76    &  69.31  &  73.57 & 77.67  \\   
         & ours  & \textbf{87.31} & \textbf{84.06} & \textbf{97.53}  & \textbf{83.43} & \textbf{88.08}  &&   \textbf{98.50} &  \textbf{68.19} &  \textbf{73.15} &  \textbf{81.53} &  \textbf{80.34} \\  
           
        \bottomrule
	\end{tabular}
	   \end{threeparttable}}\vspace{-0.1in}
    \end{center}  %\vskip -0.15in 

\end{table}

\begin{table}[t]  % \vspace{-0.1in}    
	\caption{Adversarial training performance of MoCo-v2 pre-trained ResNet-50 on CIFAR10   under the attack of PGD-10  in terms of robust  and clean accuracies.  AT-CE indicates adversarial training (AT) with CE; AT-CE-Con enhances AT-CE with the contrastive loss; AT-ours uses Core-tuning for AT.} \vspace{-0.05in}
     \label{exp_robust1}   
    \begin{center}
        \scalebox{0.67}{  
    \begin{threeparttable}  
	\begin{tabular}{lcccccccccccccccccccc}
	\toprule
    \multirow{3}{*}{Method}  &	\multicolumn{8}{c}{$\ell_2$-attack}&& \multicolumn{8}{c}{$\ell_{\infty}$-attack}\\
	 \cmidrule{2-9} \cmidrule{11-18} 
	   & \multicolumn{2}{c}{$\epsilon\small{=}0.5$} &&    \multicolumn{2}{c}{$\epsilon\small{=}1.5$} &&  
	   \multicolumn{2}{c}{$\epsilon\small{=}2.5$} && 
	   \multicolumn{2}{c}{$\epsilon\small{=}$~2/255} && 
	   \multicolumn{2}{c}{$\epsilon\small{=}$~4/255} && 
	   \multicolumn{2}{c}{$\epsilon\small{=}$~8/255}\cr  \cmidrule{2-3}    \cmidrule{5-6}  \cmidrule{8-9}  \cmidrule{11-12}  \cmidrule{14-15}  \cmidrule{17-18} 
       & Robust  &Clean  && Robust  &Clean && Robust  &Clean  && Robust  &Clean   && Robust  &Clean   && Robust  &Clean  \cr
        \midrule
        CE & 50.25 & 94.70 &&   48.29 & 94.70 &&    46.82 & 94.70   &&   25.13 & 94.70    &&12.28 &94.70    &&  4.57 & 94.70 \\    
        AT-CE & 86.59 & 92.00&&  89.60 & 94.28     &&  89.16 & 94.15  && 83.20   & 93.05          &&75.82 & 91.99  &&  69.27 & 92.79   \\    
        AT-CE-Con     & 90.74 & 94.71&&  90.29 & 94.80  &&   89.70 & 94.27    &&    85.07 & 94.56       &&79.75 & 93.79 &&  70.70 & 93.38  \\ 
        AT-ours   & \textbf{92.97} & \textbf{96.82}  && \textbf{92.32} & \textbf{96.90} && \textbf{92.05} & \textbf{96.87}  &&   \textbf{86.92} & \textbf{96.29}        && \textbf{82.01} & \textbf{95.95}    && \textbf{74.83} & \textbf{95.90}  \\  
  
        \bottomrule
        \\
        
	\end{tabular} 
\end{threeparttable}}
    \end{center}\vspace{-0.25in}
\end{table}

\subsection{Robustness to Adversarial Samples}
As is known, deep networks are fragile to adversarial attack~\cite{szegedy2013intriguing}. We  next study  whether   Core-tuning also benefits model robustness to adversarial samples in the setting of  adversarial training (AT). %wonder

\textbf{Implementation details.}  
We use MoCo-v2 pre-trained ResNet-50 as the network backbone, and
use the  Projected Gradient Descent (PGD)~\cite{madry2017towards}  to generate adversarial samples with $\ell_2$ attack (strength $\sigma\small{=}0.5$) and $\ell_{\infty}$ attack (strength $\sigma\small{=}4/255$).
During AT, we use both original samples and adversarial samples for fine-tuning. Moreover, we use the clean accuracy on original samples and the robust accuracy on adversarial samples as metrics.  More implementation details are put in Appendix C.   

\textbf{Results.}
We report the results on CIFAR10  in Table~\ref{exp_robust1} and the results on  Caltech-101, DTD and Pets in Appendix D. First, despite good clean accuracy, fine-tuning with cross-entropy cannot defend against adversarial attack, leading to poor robust accuracy.
Second, AT with cross-entropy improves the robust accuracy significantly, but it inevitably degrades the clean accuracy due to the well-known accuracy-robustness trade-off~\cite{tsipras2018robustness}. 
In contrast, the contrastive regularizer  improves both robust and clean accuracies. This is because contrastive learning helps to improve robustness generalization (\ie alleviating the distribution shifts between clean and adversarial samples). 
 Last, Core-tuning further boosts AT and, surprisingly, even achieves better clean accuracy than the standard fine-tuning with cross-entropy. 
To our knowledge, this is quite promising since even the most advanced AT methods~\cite{yang2020closer,zhang2021geometryaware} find it
difficult to overcome  the accuracy-robustness trade-off~\cite{zhang2019theoretically}. The improvement is because both contrastive learning and smooth classifier learning boost   robustness generalization.
We thus conclude that Core-tuning improves model robustness on downstream tasks.

\section{Conclusions}  \label{conclusion}
This paper studies how to fine-tune contrastive self-supervised visual models. We  theoretically show that  optimizing the contrastive loss during fine-tuning has regularization effectiveness on representation learning as well as   optimization effectiveness on classifier training, both of which benefit model fine-tuning. 
We thus propose a novel contrast-regularized tuning (Core-tuning) method  to   fine-tune CSL visual models. Promising results on   image classification and semantic segmentation verify the effectiveness of Core-tuning. Also, we empirically find that Core-tuning is beneficial to model generalization and robustness on downstream tasks. We thus recommend using Core-tuning as a standard baseline to fine-tune CSL visual models, and also call for more attention to the fine-tuning of CSL visual models on understanding its underlying theories and better approaches in the future.

\textbf{Limitation discussion.} 
One potential limitation of Core-tuning  is that it is specifically designed for and also focuses on the fine-tuning of CSL visual models. Considering the universality of Core-tuning (cf.~Table~\ref{exp_ssl_model}), we will explore the extension of  Core-tuning to better fine-tune supervised pre-trained  and other self-supervised visual models and even language models on more  tasks.  

\subsection*{Acknowledgments}
This work was partially supported by  NUS ODPRT Grant R252-000-A81-133.

% \clearpage 
% \small
% \balance
% \bibliography{reference}
% \bibliographystyle{neurips2021}

%\section*{References}
{\small
\bibliography{Main}
\bibliographystyle{plain}
}
% References follow the acknowledgments. Use unnumbered first-level heading for
% the references. Any choice of citation style is acceptable as long as you are
% consistent. It is permissible to reduce the font size to \verb+small+ (9 point)
% when listing the references.
% Note that the Reference section does not count towards the page limit.
% \medskip

% {
% \small

% [1] Alexander, J.A.\ \& Mozer, M.C.\ (1995) Template-based algorithms for
% connectionist rule extraction. In G.\ Tesauro, D.S.\ Touretzky and T.K.\ Leen
% (eds.), {\it Advances in Neural Information Processing Systems 7},
% pp.\ 609--616. Cambridge, MA: MIT Press.

% [2] Bower, J.M.\ \& Beeman, D.\ (1995) {\it The Book of GENESIS: Exploring
%   Realistic Neural Models with the GEneral NEural SImulation System.}  New York:
% TELOS/Springer--Verlag.

% [3] Hasselmo, M.E., Schnell, E.\ \& Barkai, E.\ (1995) Dynamics of learning and
% recall at excitatory recurrent synapses and cholinergic modulation in rat
% hippocampal region CA3. {\it Journal of Neuroscience} {\bf 15}(7):5249-5262.
% }

%%%%%%%%%%%%%%%%%%%%%%%%%%%%%%%%%%%%%%%%%%%%%%%%%%%%%%%%%%%%
%%%%%%%%%%%%%%%%%%%%%%%%%%%%%%%%%%%%%%%%%%%%%%%%%%%%%%%%%%%%
%%%%%%%%%%%%%%%%%%%%%%%%%%%%%%%%%%%%%%%%%%%%%%%%%%%%%%%%%%%%
%%%%%%%%%%%%%%%%%%%%%%%%%%%%%%%%%%%%%%%%%%%%%%%%%%%%%%%%%%%%
%%%%%%%%%%%%%%%%%%%%%%%%%%%%%%%%%%%%%%%%%%%%%%%%%%%%%%%%%%%%
%%%%%%%%%%%%%%%%%%%%%%%%%%%%%%%%%%%%%%%%%%%%%%%%%%%%%%%%%%%%
%%%%%%%%%%%%%%%%%%%%%%%%%%%%%%%%%%%%%%%%%%%%%%%%%%%%%%%%%%%%

\clearpage

\appendix

\begin{table}
	\setlength{\tabcolsep}{0.2cm}
	\begin{tabular}{p{0.97\columnwidth}}
		\nipstophline 
		\vspace{-2pt}
		\centering
		\textbf{\Large{Supplementary Materials:\\Unleashing the Power of Contrastive Self-Supervised Visual Models via Contrast-Regularized Fine-Tuning}}
		\vspace{-5pt}
		\nipsbottomhline   
	\end{tabular}
\end{table}\vspace{-5pt}

This supplementary material provides proofs for the analysis of the contrastive loss (cf.~Appendix A), the pseudo-code of the proposed method (cf.~Appendix B), more implementation details (cf.~Appendix C), and more empirical results and analysis (cf.~Appendices D and E).
% \begin{itemize}
%     \item Appendix A provides proofs for the analysis of the contrastive loss.
%     \item Appendix B provides the pseudo-code of the proposed method.
%     \item Appendix C provides more implementation details of the proposed method.
%     \item Appendix D provides  more empirical results of the proposed method.
%     \item Appendix E provides  more analysis for our proposed method.
% \end{itemize}
% 

\section{Proof of Theoretical Analysis}  
This appendix provides proofs for both Theorems 1 and 2.
\subsection{Proof for Theorem 1}  
% \begin{thm}\label{propo1}
% Assuming the features are $\ell_2$-normalized and   the classes are balanced with equal data number, minimizing the contrastive loss is equivalent  to  minimizing the class-conditional entropy $\mathcal{H}(Z|Y)$ and maximizing the feature entropy $\mathcal{H}(Z)$: 
% \begin{align}
%   \mathcal{L}_{con} ~    \propto ~~    \mathcal{H}(Z|Y) ~ - ~\mathcal{H}(Z)  \nonumber
% \end{align}   
% \end{thm}

\begin{proof}
We follow the notations in the main paper and further denote the sample set of the class $k$ by $\mathcal{Z}_k$. Moreover, we assume the classes of samples are balanced so that the sample number of each class is constant $|\mathcal{Z}_k|=\frac{n}{K}$, where $n$ denotes the total  number of samples and $K$ indicates the  number of classes. Let us start by splitting the contrastive loss into two terms.
\begin{align}\label{eq_contrastive}
    \mathcal{L}_{con} &\small{=-}\frac{1}{n}\sum_{i=1}^n \frac{1}{|P_i|}\sum_{z_j\in P_i}  \log \frac{e^{(v_i^{\top} v_j/\tau)}}{\sum_{v_k\small{\in} A_i}e^{(v_i^{\top} v_k/\tau)}} \nonumber \\
    & \small{=} \small{-}\frac{1}{n}\sum_{i=1}^n \frac{1}{|P_i|}\sum_{z_j\in P_i} \frac{z_i^{\top}z_j}{\tau}\small{+} \frac{1}{n} \sum_{i=1}^n \log \sum_{z_k\in A_i} e^{(\frac{z_i^{\top}z_k}{\tau})}. 
\end{align}
Let $c_k \small{=} \frac{1}{|\mathcal{Z}_k|}\sum_{z\in\mathcal{Z}_k}z$ denote the hard mean of all features from the class $k$,  and let the symbol $\small{\overset{c}{=}}$ indicate  equality up to a multiplicative and/or additive constant. We first analyze the first term in Eq.~(\ref{eq_contrastive}) by connecting it to a tightness term  of the center loss, \ie $\sum_{z_i\in \mathcal{Z}_k}\|z_i\small{-}c_k\|^2$~\cite{wen2016discriminative}:
\begin{align}
     \sum_{z_i, z_j\in \mathcal{Z}_k} -\frac{z_i^{\top}z_j}{\tau} & \overset{c}{=}  \frac{1}{|\mathcal{Z}_k|} \sum_{z_i, z_j\in \mathcal{Z}_k} - z_i^{\top}z_j  \nonumber \\
     & \overset{c}{=}  \frac{1}{|\mathcal{Z}_k|} \sum_{z_i, z_j\in \mathcal{Z}_k} \|z_i\|^2 - z_i^{\top}z_j   \nonumber \\
     %&  = \frac{1}{2} \left[\sum_{z_i\in \mathcal{Z}_k} \|z_i\|^2 \small{+} \sum_{z_j\in \mathcal{Z}_k} \|z_j\|^2\right] \nonumber  \\ &~~~~~~-~~  \frac{1}{2|\mathcal{Z}_k|} \frac{1}{|\mathcal{Z}_k|} \sum_{z_i\in \mathcal{Z}_k} \sum_{z_j\in \mathcal{Z}_k}    2z_i^{\top}z_j \nonumber \\
     &=  \sum_{z_i\in \mathcal{Z}_k} \|z_i\|^2 -  \frac{1}{|\mathcal{Z}_k|} \sum_{z_i\in \mathcal{Z}_k} \sum_{z_j\in \mathcal{Z}_k}    z_i^{\top}z_j \nonumber \\
    &  = \sum_{z_i\in \mathcal{Z}_k} \|z_i\|^2 - 2 \frac{1}{|\mathcal{Z}_k|} \sum_{z_i\in \mathcal{Z}_k} \sum_{z_j\in \mathcal{Z}_k}    z_i^{\top}z_j \nonumber \\  &~~~~~~~~ +   \frac{1}{|\mathcal{Z}_k|} \sum_{z_i\in \mathcal{Z}_k} \sum_{z_j\in \mathcal{Z}_k}    z_i^{\top}z_j \nonumber \\
     &=  \sum_{z_i\in \mathcal{Z}_k} \|z_i\|^2 - 2 z_i^{\top}c_k  +   \|c_k\|^2\nonumber \\
      &=\sum_{z_i\in \mathcal{Z}_k}\|z_i\small{-}c_k\|^2,  \nonumber  
\end{align}
where we use the property of $\ell_2$-normalized features that $\|z_i\|^2 \small{=}  \|z_j\|^2\small{=} 1$ and the definition of the class hard mean $c_k \small{=} \frac{1}{|\mathcal{Z}_k|}\sum_{z\in\mathcal{Z}_k}z$.

By summing over all classes $k$, we obtain:
\begin{align}
    \sum_{i=1}^n\sum_{z_j\in P_i} -\frac{z_i^{\top}z_j}{\tau} ~~ {\overset{c}{=}} ~~ \sum_{i=1}^n \|z_i\small{-}c_{y_i}\|^2. \nonumber
\end{align}

Based on this equation, following~\cite{boudiaf2020unifying}, we can interpret the first term in Eq.~(\ref{eq_contrastive}) as a conditional cross-entropy between $Z$ and another random variable $\bar{Z}$, whose conditional distribution given $Y$ is a standard Gaussian centered around $c_Y\small{:}\bar{Z}|Y\small{\sim}\mathcal{N}(c_y,i)$:
\begin{align}
    - \frac{1}{n}\sum_{i=1}^n \frac{1}{|P_i|}\sum_{z_j\in P_i}  \frac{z_i^{\top}z_j }{\tau}   \overset{c}{=}   \mathcal{H}(Z;\bar{Z}|Y) = \mathcal{H}(Z|Y) \small{+}\mathcal{D}_{KL}(Z||\bar{Z}|Y). \nonumber
\end{align}
Based on this, we know that the first term in  Eq.~(\ref{eq_contrastive}) is an upper bound on the conditional entropy of features $Z$ given labels $Y$:
\begin{align}%\frac{z_i^{\top}z_j}{\tau}
  - \frac{1}{n}\sum_{i=1}^n \frac{1}{|P_i|}\sum_{z_j\in P_i} \frac{z_i^{\top}z_j}{\tau}  \overset{c}{\geq} \mathcal{H}(Z|Y).   \nonumber
\end{align}
%When $\bar{Z}|Y\small{\sim}\mathcal{N}(c_y,i)$, this bound is tight~\cite{boudiaf2020unifying}. 
where the symbol $\small{\overset{c}{\geq}}$ indicates  ``larger than" up to a multiplicative and/or an additive constant. When $Z|Y\small{\sim}\mathcal{N}(c_y,i)$, the bound is tight.  As a result, minimizing the first term in  Eq.~(\ref{eq_contrastive})   is equivalent  to minimizing $\mathcal{H}(Z|Y)$:
\begin{align}\label{eq_first_term}
  - \frac{1}{n}\sum_{i=1}^n \frac{1}{|P_i|}\sum_{z_j\in P_i} \frac{z_i^{\top}z_j}{\tau} \propto \mathcal{H}(Z|Y).   
\end{align}
This concludes the proof for the relationship of the first term in Eq.~(\ref{eq_contrastive}).

We then analyze the second term  in Eq.~(\ref{eq_contrastive}), which has the following relationship:
\begin{align}\label{eq2}
    &\frac{1}{n} \sum_{i=1}^n \log \sum_{z_k\in A_i} e^{(\frac{z_i^{\top}z_k}{\tau})} \nonumber \\ 
    & =  \frac{1}{n} \sum_{i=1}^n \log \left(\sum_{k:y_i=y_k} e^{(\frac{z_i^{\top}z_k}{\tau})} \small{+} \sum_{k:y_i \ne y_k} e^{(\frac{z_i^{\top}z_k}{\tau})}\right)   \nonumber \\ 
    & \geq \frac{1}{n} \sum_{i=1}^n \log \left(\sum_{k:y_i \ne y_k} e^{(\frac{z_i^{\top}z_k}{\tau})}\right)    \nonumber \\ 
    & \overset{c}{\geq} \frac{1}{n} \sum_{i=1}^n \sum_{k:y_i \ne y_k} \frac{z_i^{\top}z_k}{\tau}  \nonumber \\ 
    & = \frac{1}{n} \sum_{i=1}^n \sum_{k=1}^n  \frac{z_i^{\top}z_k}{\tau}  - \frac{1}{n} \sum_{i=1}^n \sum_{k:y_i = y_k} \frac{z_i^{\top}z_k}{\tau}  \nonumber \\ 
    & \overset{c}{=} - \frac{1}{n} \sum_{i=1}^n \sum_{k=1}^n \|z_i \small{-}z_k\|^2  - \frac{1}{n} \sum_{i=1}^n \sum_{k:y_i = y_k} \frac{z_i^{\top}z_k}{\tau},   
\end{align}
%bh: Jenson -> Jensen
%bh: remove 'the'
where we use Jensen's inequality in the fourth line. 
%the concavity of $x\small{\xrightarrow{}\log(x)}$ and 
The first term   in Eq.~(\ref{eq2}) is close to the differential entropy estimator of features $Z$ provided by~\cite{wang2011information}:
\begin{align}\label{eq3}
\hat{\mathcal{H}}(Z)   = \frac{d}{n(n-1)}  \sum_{i=1}^n \sum_{k=1}^n \log \|z_i \small{-}z_k\|^2  
  \overset{c}{=} \frac{1}{n} \sum_{i=1}^n \sum_{k=1}^n  \log \|z_i \small{-}z_k\|^2 \propto \frac{1}{n} \sum_{i=1}^n \sum_{k=1}^n   \|z_i \small{-}z_k\|^2,
\end{align}
where $d$ is the dimension of features. Combining Eq.~(\ref{eq2}) and Eq.~(\ref{eq3}) leads to:
\begin{align} \label{eq4}
    \frac{1}{n} \sum_{i=1}^n \log \sum_{z_k\in A_i} e^{(\frac{z_i^{\top}z_k}{\tau})}  \small{\overset{c}{\geq}} \small{-}\mathcal{H}(Z) \small{-} \frac{1}{n} \sum_{i=1}^n \sum_{k:y_i = y_k} \frac{z_i^{\top}z_k}{\tau}.    
\end{align}

%\newpage
The second term in the right side of Eq.~(\ref{eq4}) is essentially a redundant term with the first term in Eq.~(\ref{eq_contrastive}), so we ignore it here. Then, we know that minimizing the second term in Eq.~(\ref{eq_contrastive}) is equivalent to maximizing  $\mathcal{H}(Z)$:
\begin{align} \label{eq_second_term}
    \frac{1}{n} \sum_{i=1}^n \log \sum_{z_k\in A_i} e^{(\frac{z_i^{\top}z_k}{\tau})} \propto \small{-}\mathcal{H}(Z).      
\end{align}
Combining Eq.~(\ref{eq_first_term}) and Eq.~(\ref{eq_second_term}), we conclude the proof of Theorem~1.
\end{proof}

\subsection{Proof for Theorem 2}  
% \begin{thm} \label{thm1}
% Assuming the features are $\ell_2$-normalized and  the classes are balanced, the contrastive loss  is positive proportional  to  the infimum of conditional cross-entropy $\mathcal{H}(Y;\hat{Y}|Z)$, where the infimum is taken over classifiers: 
% \begin{align}
%     \mathcal{L}_{con}~\propto~ \inf  \underbrace{\mathcal{H}(Y;\hat{Y}|Z)}_{\text{Conditional CE}}  ~ - ~   \mathcal{H}(Y) \nonumber
% \end{align}
% \end{thm} 

\begin{proof}
The mutual information between  features $Z$ and  labels $Y$ can be defined in two ways:
\begin{align}\label{eq_theorem_proof1}
    \mathcal{I}(Z;Y)=\mathcal{H}(Y)-\mathcal{H}(Y|Z) =  \mathcal{H}(Z)-\mathcal{H}(Z|Y).
\end{align}
Based on Theorem~1, we know that:
\begin{align}\label{eq_theorem_proof2}
    \mathcal{L}_{con} \propto \mathcal{H}(Z|Y) - \mathcal{H}(Z) = -\mathcal{I}(Z;Y).
\end{align}
Combining Eq.~(\ref{eq_theorem_proof1}) and Eq.~(\ref{eq_theorem_proof2}), we have:
\begin{align}\label{eq_theorem_proof3}
    \mathcal{L}_{con} \propto \mathcal{H}(Y|Z) -\mathcal{H}(Y).
\end{align}
Then, we relate the conditional entropy $\mathcal{H}(Y|Z)$ to the  cross entropy loss:
\begin{align}\label{eq_theorem_proof4}
  \mathcal{H}(Y;\hat{Y}|Z)   = \mathcal{H}(Y|Z) +\mathcal{D}_{KL}(Y\|\hat{Y}|Z).
\end{align}
According to  Eq.~(\ref{eq_theorem_proof4}), when we  minimize cross-entropy $\mathcal{H}(Y;\hat{Y}|Z)$,  we implicitly minimize both  $\mathcal{H}(Y|Z)$ and $\mathcal{D}_{KL}(Y\|\hat{Y}|Z)$. In fact, the optimization could be decoupled into 2 steps in a maximize-minimize (or bound-optimization) way~\cite{boudiaf2020unifying}. The first step fixes the parameters of the network encoder  and only minimizes Eq.~(\ref{eq_theorem_proof4}) with respect to the parameters of the network classifier. As this  step, $\mathcal{H}(Y|Z)$ is fixed  and the predictions $\hat{Y}$ are adjusted  to minimize $\mathcal{D}_{KL}(Y\|\hat{Y}|Z)$. Ideally, $\mathcal{D}_{KL}(Y\|\hat{Y}|Z)$ would vanish at the end of this step~\cite{boudiaf2020unifying}. In this sense, we know that:
\begin{align}\label{eq_theorem_proof5}
 \mathcal{H}(Y|Z) = \inf  \mathcal{H}(Y;\hat{Y}|Z).
\end{align}
The second step fixes the classifier and  minimizes Eq.~(\ref{eq_theorem_proof4}) with respect to  the encoder. By combining Eq.~(\ref{eq_theorem_proof3}) and Eq.~(\ref{eq_theorem_proof5}), we conclude the proof of Theorem~2.
\end{proof}

\section{Pseudo-code of Core-tuning}  
We summarize the scheme of Core-tuning in Algorithm~\ref{alg:overall}.  Here, all hard pair generation is conducted within each sample batch.
%\vspace{-0.1in}
\begin{algorithm}[H]
\small
\caption{The training scheme of Core-tuning.}
\label{alg:overall}
\begin{algorithmic}[1]
\REQUIRE{Pre-trained encoder $G_e$; Loss factor $\eta$; Mixup factor $\alpha$; Batch size $B$; Epoch number $T$.}
\ENSURE{Classifier $G_y$; Projection head $G_c$.} 
\FOR{t=1,...,T}
\STATE Sample a batch of training data $\{(x_i,y_i)\}_{i=1}^B$;
\STATE Obtain  features $z_i = G_e(x_i)$ for each sample;
\FOR{i=1,...,B}
\STATE Construct  positive pair set $P_i$ and  full pair set $A_i$ for $z_i$;
\STATE Generate  hard positive pair $(z_i^+,y_i^+)$ and add it to $P_i$, $A_i$;  
\STATE Generate  hard negative pair $(z_i^-,y_i^-)$ and add it to  $A_i$;  
\ENDFOR 
\STATE Obtain contrastive features $v_i = G_c(z_i)$ for all features;  %~~~ // original features and all generated features
\STATE Compute the focal contrastive loss $\mathcal{L}_{con}^f$;
\STATE Predict $\hat{y}_i\small{=}G_y(z_i)$ for the original  and  generated samples;  
\STATE Compute the cross-entropy loss $\mathcal{L}_{ce}^m$;
\STATE loss.backward();  ~~~ // loss $=\mathcal{L}_{ce}^m + \eta \mathcal{L}_{con}^f$.
\ENDFOR 
\end{algorithmic} 
\end{algorithm}%\vspace{-0.2in}
 %; Threshold $\lambda_p\small{=}\lambda_n\small{=}0.8$
 
 \newpage
\section{More Experimental Details}  
 
\subsection{Implementation Details of Feature Visualization}  

In the feature visualization, we train ResNet-18 on CIFAR10 with two kinds of losses, \ie (1) cross-entropy $\mathcal{L}_{ce}$; (2)  cross-entropy and the contrastive loss $\mathcal{L}_{ce}\small{+}\mathcal{L}_{con}$. For better visualization,  following~\cite{liu2017sphereface}, we add two fully connected layers before the classifier. The two layers first map the 512-dimensional features  to  a  3-dimensional feature sphere  and then map back to the 10-dimensional 
feature space for prediction. The contrastive loss $\mathcal{L}_{con}$ is  enforced on the 3-dimensional features. After training, we visualize the  3-dimensional features learned by ResNet-18  in MATLAB. 

\subsection{More Details of Image Classification}   
\textbf{Dataset details.}   
Following~\cite{kornblith2019better}, we test on 9 natural image datasets, including ImageNet20 (a subset of ImageNet with 20 classes)~\cite{deng2009imagenet}, CIFAR10, CIFAR100~\cite{krizhevsky2009learning}, Caltech-101~\cite{fei2004learning}, DTD~\cite{cimpoi2014describing}, FGVC Aircraft~\cite{maji2013fine}, Standard Cars~\cite{krausecollecting}, Oxford-IIIT Pets~\cite{parkhi2012cats} and Oxford 102 Flowers~\cite{nilsback2008automated}. 
In addition, considering real-world datasets may  be class-imbalanced~\cite{zhang2021test, zhang2021deep, zhang2018online,zhao2018adaptive}, we also evaluate Core-tuning on the iNaturalist18 dataset~\cite{van2018inaturalist}.
Most datasets are obtained from their official websites, except ImageNet20 and Oxford 102 Flowers. The ImageNet20 dataset is obtained by combining two open-source ImageNet subsets with 10 classes, \ie  ImaegNette and ImageWoof~\cite{imagewang}. Moreover, Oxford 102 Flowers is obtained from Kaggle\footnote{\url{https://www.kaggle.com/c/oxford-102-flower-pytorch}.}. These datasets cover a wide range of classification tasks, including coarse-grained object classification (\ie ImageNet20, CIFAR, Caltech-101), fine-grained object classification (\ie Cars, Aircraft, Pets) and texture classification (\ie DTD). The statistics of all datasets are reported in Table~\ref{dataset}.

%bh: in -> from
%\blue{iNaturalist18 and hyper-parameters}. 

\begin{table}[h]   \vspace{-0.05in}
    \caption{Statistics of datasets.}\label{dataset}  
    \begin{center}\vspace{-0.1in}
    \scalebox{0.68}{  
    \begin{threeparttable} 
	\begin{tabular}{lccc}\toprule
        DataSet & $\#$Classes &  $\#$ Training &  $\#$ Test  \\ \midrule    
 
        ImageNet20~\cite{imagewang,deng2009imagenet} 	& 20 & 18,494	& 7,854  \\
        CIFAR10~\cite{krizhevsky2009learning} & 10 & 50,000 & 10,000 \\
        CIFAR100~\cite{krizhevsky2009learning} & 100 & 50,000 & 10,000 \\
        Caltech-101~\cite{fei2004learning} & 102 & 3,060 & 6,084 \\ 
        Describable~Textures (DTD)~\cite{cimpoi2014describing}   & 47 & 3,760 & 1,880 \\
        FGVG Aircraft~\cite{maji2013fine} & 100 &6,667 &3,333  \\ 
        
        Standard Cars~\cite{krausecollecting} & 196 & 8,144 & 8,041 \\      
        Oxford-IIIT Pets~\cite{parkhi2012cats} & 37 & 3,680 & 3,369\\ 
        Oxford 102 Flowers~\cite{nilsback2008automated} & 102 & 6,552 & 818 \\ \midrule
        iNaturalist18~\cite{van2018inaturalist} & 8,142 & 437,513 & 24,426 \\
        \bottomrule
	\end{tabular}  
    \end{threeparttable}}
    \end{center} \vspace{-0.1in}
\end{table}
 
\textbf{Implementation details.}
We implement all methods in PyTorch. All checkpoints of self-supervised  models are  provided by the authors or by the PyContrast GitHub repository\footnote{\url{https://github.com/HobbitLong/PyContrast}.}.  
For most datasets, following~\cite{chen2020simple, kornblith2019better},  we preprocess images  via random resized  crops  to $224\small{\times}224$ and flips.  At the test time, we resize images to   $256\small{\times}256$ and then take a $224\small{\times}224$ center crop. In such a  setting, however, we find it difficult to reproduce the  performance of some CSL models~\cite{chen2020simple}. Therefore,  for some datasets (\eg CIFAR10 and Aircraft), we   resize images to different scales and use rotation augmentations. Although the preprocessing of some datasets is slightly different from~\cite{chen2020simple}, the results in this paper are obtained with the same preprocessing method \wrt each dataset and thus are fair.

Following~\cite{kornblith2019better}, we initialize networks with the checkpoints of  contrastive self-supervised  models.
For most datasets, we fine-tune networks for 100 epochs using Nesterov momentum via the cosine learning rate schedule. For ImageNet20, we fine-tune networks using stochastic gradient descent via the linear learning rate decay. For iNaturalist18, we fine-tune networks for 160 epochs.  For all datasets, the momentum parameter is set to 0.9, while the   factor of weight decay is set to $10^{-4}$.
As for  Core-tuning, we set the clipping thresholds of hard negative generation to be $\lambda_n\small{=}0.8$ and the temperature $\tau\small{=}0.07$. The dimension of the contrastive features is 256 and the depth of non-linear projection is 2 layers. 
Following~\cite{chen2020simple}, we perform hyper-parameter tuning for each dataset. Specifically, we select the batch size from $\{64,128,256\}$, the initial learning rate from $\{0.01,0.1\}$ and $\eta/\alpha$ from $\{0.1,1,10\}$.  
The experiments are conducted on 4 TITAN RTX 2080 GPUs for iNaturalist18, and 1 GPU for all other datasets.
All results are averaged over 3 runs. We adopt the top-1 accuracy as the metric. The statistics of the used hyper-parameters   are provided in Table~\ref{statistic_parameter}. For other baselines, we use the same training setting for each dataset, and tune their hyper-parameters as best as possible.

\begin{table*}[t] 
\vspace{-0.1in}
	\caption{Statistics of the used hyper-parameters in Core-tuning.} 
     \label{statistic_parameter} 
    \begin{center}
     \scalebox{0.68}{  
    \begin{threeparttable} 
	\begin{tabular}{l|c|c|c|c|c|c|c|c|c|c}\hline
        Hyper-parameter  &   ImageNet20 & CIFAR10 & CIFAR100 & Caltech101 & DTD  &Aircraft& Cars & Pets & Flowers  & iNarutalist18 \\ \hline
        epochs &  \multicolumn{9}{c|}{100} & 160 \\     \hline
        batch size  & 256  & 256 & 256  & 256  &   256 &  64  &64  & 64  & 64   & 128 \\  \hline                
        loss trade-off factor $\eta$ & 0.1  & 0.1 &    1  & 1   & 0.1   &0.1  & 0.1 & 0.1 &  1  & 10  \\\hline
        mixup factor $\alpha$  & 1 & 1  &0.1 & 0.1 & 1    & 0.1  & 0.1    & 1  & 0.1 & 1 \\  \hline
        
        learning rate (lr)    &  0.1 & 0.01 &0.01 &  0.01& 0.01  & 0.01 & 0.01 &   0.01 &  0.01  &  0.1 \\    \hline
        lr schedule    & linear & \multicolumn{9}{c}{cosine decay}\\  \hline

        temperature $\tau$    & \multicolumn{10}{c}{0.07}\\  \hline
        threshold $\lambda_n$  &  \multicolumn{10}{c}{0.8}\\      \hline    
        weight decay factor  & \multicolumn{10}{c}{$10^{-4}$}\\         \hline 
        momentum factor&  \multicolumn{10}{c}{0.9}\\  \hline
        projection dimension& \multicolumn{10}{c}{256}\\  \hline
        projection depth&    \multicolumn{10}{c}{2 layers}\\  
        \hline
	\end{tabular} 
    \end{threeparttable}} 
    \end{center}  
  \vspace{-0.15in}
\end{table*}

\subsection{More Details of Domain Generalization}   

\textbf{Dataset details.}  
We use 3 benchmark datasets, \ie PACS~\cite{li2017deeper}, VLCS~\cite{fang2013unbiased}  and Office-Home~\cite{venkateswara2017deep}.  The data statistics are shown in Table~\ref{dataset_DG}, where each dataset has 4 domains. In each  setting, we select 3  domains to fine-tune the networks and then test on the rest of the unseen domains. The key challenge is the distribution discrepancies among domains, leading to poor performance of neural networks on the target domain~\cite{zhang2019whole,zhang2020collaborative}.
%bh: we select 3 domains to fine-tune the networks and test them on the rest of the unseen domains.

\begin{table}[h]  
%\vspace{-0.2in}
	\caption{Statistics of datasets.}\label{dataset_DG}  
  \vspace{-0.1in}
    \begin{center}
    \scalebox{0.8}{ 
    \begin{threeparttable}  
	\begin{tabular}{lcccc}\toprule
        DataSet  & $\#$Domains & $\#$Classes &  $\#$Samples &  Size of images\\ \midrule    
        PACS & 4          & 7	& 9,991	 & (3,224,224)	  \\  
        VLCS &  4   	& 5 	& 10,729 & (3,224,224)		\\ 
        Office-Home &  4   	& 65 	& 15,588 & (3,224,224)	\\	   
        \bottomrule
	\end{tabular}
    \end{threeparttable}}
    \end{center} \vspace{-0.1in}
\end{table}

\begin{figure*}[h]  
 \centerline{\includegraphics[width=15cm]{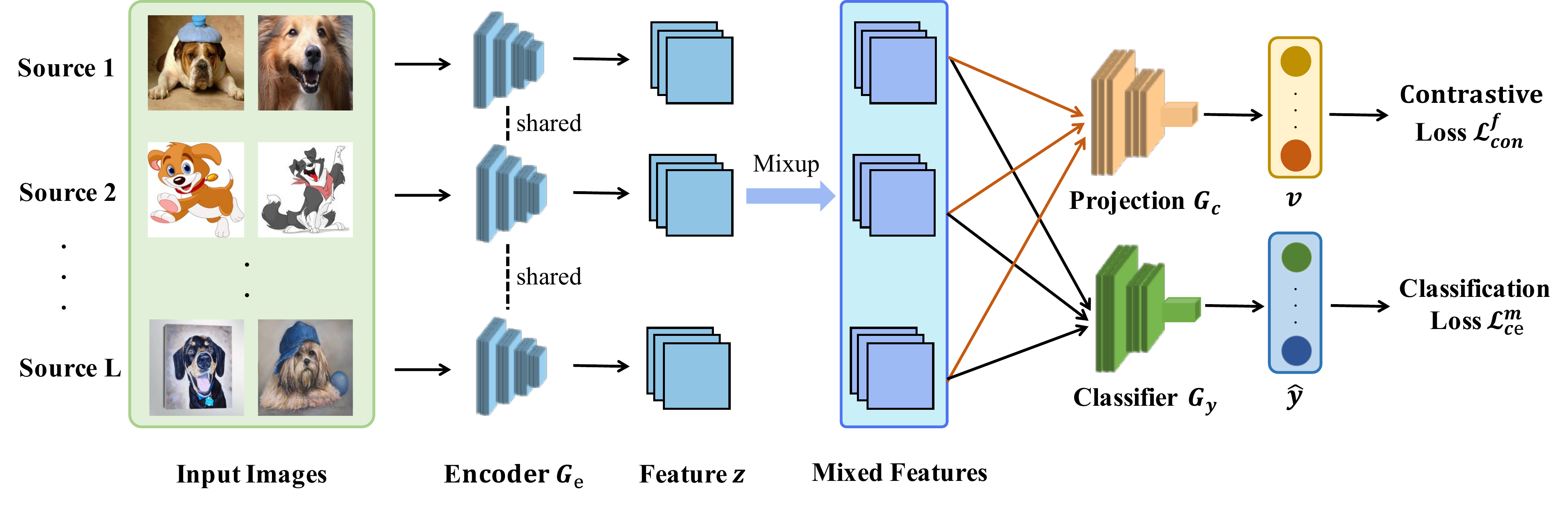}} % \vspace{-0.1in}
 \caption{The overall scheme of Core-tuning in the setting of cross-domain generalization.}\vspace{-0.1in}
 \label{DG_framework}  % \vspace{-0.1in} Best viewed in color.
\end{figure*}

\textbf{Implementation details.} 
The overall scheme of Core-tuning for domain generalization is shown in Figure~\ref{DG_framework}.
The experiments are implemented based on  the DomainBed repository~\cite{gulrajani2020search} in PyTorch.  
During fine-tuning, we preprocess images through random resized crops to $224\small{\times}224$, horizon flips, color jitter and random gray scale. At the test time, we directly resize images to $224\small{\times}224$. 
We initialize ResNet-50 with the weights of the MoCo-v2 pre-trained model, and fine-tune it for 20,000 steps at a batch size of 32 using the Adam optimizer on a single TITAN RTX 2080 GPU. We set the initial learning rate to  $5\small{\times}10^{-5}$ and adjust it via the exponential learning rate decay. All other hyper-parameters of Core-tuning are the same as image classification. Besides, we use Accuracy as the metric  in domain generalization. 
 
\subsection{Implementation Details of Robustness Training}   
We conduct this experiment in PyTorch. We take Caltech-101, DTD, Pets, and CIFAR10 as datasets, whose preprocessing are the same as  the ones in image classification.
We use MoCo-v2 pre-trained ResNet-50 as the backbone, and
use   Projected Gradient Descent (PGD)~\cite{madry2017towards}  to generate adversarial samples.
During adversarial training (AT), we use both clean and adversarial samples for training with various fine-tuning methods on a single TITAN RTX 2080 GPU.  Other training schemes (\eg the optimizer, the hyper-parameters, the learning rate scheme) are the   same as   image classification.

\newpage
\section{More Experimental Results}  

\subsection{More Results on Domain Generalization}

This appendix further reports the results of domain generalization on OfficeHome. The observations from Table~\ref{exp_generalization1} are same to the main text. First,  when fine-tuning with cross-entropy, the contrastive self-supervised model  performs  worse than the supervised pre-trained model. This results from the relatively worse discriminative abilities of the contrastive self-supervised model, which can also be found in  Table~\ref{exp_classification}. Second, enforcing contrastive regularizer during fine-tuning improves domain generalization performance, since the contrastive  regularizer helps to learn more discriminative features (cf.~Theorem~\ref{thm0}) and also helps to alleviate distribution shifts among  domains~\cite{kang2019contrastive}, hence leading to better performance. Last, Core-tuning further improves the generalization performance of models on all datasets. This is because hard pair generation further boosts contrastive learning,  while smooth classifier learning   also benefits model generalizability. We thus conclude that Core-tuning improves model generalization on downstream tasks.

\begin{table}[h]    
\caption{Domain generalization accuracies of various fine-tuning methods for MoCo-v2 pre-trained ResNet-50 the on Office-Home dataset. CE means cross-entropy;  CE-Con enhances CE with the contrastive loss. Here, A/C/P/R are four domains in Office-Home.}  
    \label{exp_generalization1} 
    \begin{center}
    \scalebox{0.73}{  
    \begin{threeparttable} 
	\begin{tabular}{ccccccc}\toprule
 
        \multirow{2}{*}{Pre-training}  &  \multirow{2}{*}{Fine-tuning}  &      \multicolumn{5}{c}{Office-Home} \cr  \cmidrule{3-7} 
        && A   & C  & P   &  R & Avg.\\\toprule 
        Supervised &  CE &  56.08     & 50.83    & 72.49 &  75.21 & 63.82  \\   \midrule  
        \multirow{3}{*}{MoCo-v2} & CE&     50.31     & 48.91  & 64.72  & 68.76 & 58.18 \\    
         & CE-Con&  55.87 & 50.23 & 71.51 & 74.99 & 63.15  \\   
        & ours &  \textbf{58.70} &  \textbf{52.43} &  \textbf{72.89} &  \textbf{75.36} &  \textbf{64.85}      \\ 
          
        \bottomrule
	\end{tabular}
	   \end{threeparttable}}
    \end{center}  

\end{table}

\subsection{More Results on Adversarial Training}  
In the main paper, we apply Core-tuning to adversarial training on CIFAR10, while this appendix further provides the results of  adversarial training on three other natural image datasets, \ie Caltech-101, DTD and Pets. 
We draw several observations based on the results on 3 image datasets  in Table~\ref{exp_robust}. First, despite good clean accuracy, standard fine-tuning with cross-entropy cannot defend against adversarial attack, leading to poor robust accuracy.
Second, AT with cross-entropy improves the robust accuracy significantly, but it inevitably degrades the clean accuracy due to the accuracy-robustness trade-off~\cite{tsipras2018robustness}. 
In contrast, the contrastive regularizer  improves both robust and clean accuracies. This is because contrastive learning helps to improve robustness generalization (\ie alleviating the distribution shifts between clean samples and adversarial samples), thus leading to better performance. 
 Last, Core-tuning further boosts AT and, surprisingly, even achieves better clean accuracy than the standard fine-tuning under the $\ell_2$ attack. 
To our knowledge, this is quite promising since even the most advanced AT methods~\cite{yang2020closer,zhang2021geometryaware} find it
difficult to conquer  the accuracy-robustness trade-off~\cite{zhang2019theoretically}. The improvement is mainly derived from that both contrastive learning and smooth classifier learning boost the robustness generalization.
We thus conclude that Core-tuning is beneficial to  model robustness. We also hope that Core-tuning  can motivate people to rethink the accuracy-robustness trade-off in adversarial training in the future.

  \begin{table}[h]  
 %   \vskip -0.1in 
	\caption{Adversarial training performance of MoCo-v2 pre-trained ResNet-50 under the attack of PGD-10 in terms of robust  and clean accuracies. CE indicates cross-entropy; AT-CE indicates adversarial training (AT) with CE; AT-CE-Con enhances AT-CE with the contrastive loss; AT-ours uses Core-tuning for AT.}
     \label{exp_robust} 
    \begin{center}
        \scalebox{0.68}{  
    \begin{threeparttable}  
	\begin{tabular}{lccccccccccccccccc}
	\toprule
	\multirow{3}{*}{Method}   & \multicolumn{8}{c}{PGD - $\ell_2$ attack ($\epsilon =0.5$)}  && \multicolumn{8}{c}{PGD - $\ell_{\infty}$ attack ($\epsilon=$~4/255)} \\
	\cmidrule{2-9}  \cmidrule{11-18}  
	   &  \multicolumn{2}{c}{Caltech101} &&  \multicolumn{2}{c}{DTD} &&  \multicolumn{2}{c}{Pets} &&  \multicolumn{2}{c}{Caltech101} &&  \multicolumn{2}{c}{DTD} &&  \multicolumn{2}{c}{Pets} \cr  \cmidrule{2-3}    \cmidrule{5-6} \cmidrule{8-9}   \cmidrule{11-12} \cmidrule{14-15} \cmidrule{17-18}
      & Robust  &Clean && Robust  &Clean  &&Robust  &Clean &&   Robust  &Clean && Robust  &Clean  &&Robust  &Clean\cr
      \toprule
        CE & 55.69 & 91.87    &&42.25 & 71.68  && 30.94 & 89.05 && 27.03 & \textbf{91.87}   && 18.37& \textbf{71.68}  && 4.63 & \textbf{89.05} \\       
        AT-CE & 87.35 & 91.61   && 61.93 & 68.81 && 78.67 & 86.25 && 78.61   &90.65    && 47.27 & 67.13 && 63.59 & 84.21 \\      
        AT-CE-Con     & 88.67 & 92.61  && 64.75 &71.24 && 79.53 & 87.01 &&79.87 & 91.08  &&48.95  &69.07  && 65.60& 86.85\\ 
        AT-ours   & \textbf{89.21} & \textbf{92.83}   && \textbf{66.49} & \textbf{72.94}  && \textbf{82.54} & \textbf{89.22} && \textbf{80.73} & 91.64   &&\textbf{49.43} & 70.65  && \textbf{67.98}& 87.20\\  
 
        \bottomrule 
        
	\end{tabular} 
\end{threeparttable}}
    \end{center} % \vskip -0.2in 
\end{table}

\subsection{More Results  on Image Classification}
\textbf{The results with standard errors.} In the main paper, we report the results of image classification and ablations studies on 9 natural image datasets in terms of the average accuracy.  To make the results more complete, this appendix further reports the results with their standard errors (cf.~Tables~\ref{exp_classification1}-\ref{exp_ablation1}).

\begin{table*}[h] 
  \vspace{-0.05in}
	\caption{Comparisons of various fine-tuning methods for MoCo-v2 pre-trained ResNet-50  on image classification in terms of top-1 accuracy. Here, ``Avg.'' indicates the average accuracy over 9 datasets.  SL-CE-tuning denotes supervised pre-training on ImageNet and then fine-tuning with cross-entropy.} \label{exp_classification1}   % explain avg.
	 %  \vspace{0.1in}
    \begin{center}
    \scalebox{0.75}{ 
    \begin{threeparttable}  
	\begin{tabular}{lccccc}\toprule
        Algorithm  & ImageNet20 & CIFAR10 & CIFAR100 & Caltech101 & DTD    \\ \toprule   
        %SL-CE & 94.23+/-0.07 & 83.40+/-0.12& 93.65+/-0.21 & 74.40+/-0.45 \\  
         
        SL-CE-tuning & 91.01+/-1.27 &94.23+/-0.07 & 83.40+/-0.12 & 93.65+/-0.21 & 74.40+/-0.45 \\ \midrule
        CE-tuning & 88.28+/-0.47  & 94.70+/-0.39 &80.27+/-0.60 & 91.87+/-0.18  & 71.68+/-0.53  \\    
        L2SP~\cite{li2018explicit}  &88.49+/-0.40 &95.14+/-0.22 & 81.43+/-0.22 &  91.98+/-0.07 & 72.18+/-0.61 \\  
        M$\&$M~\cite{zhan2018mix}    & 88.53+/-0.21 & 95.02+/-0.07 & 80.58+/-0.19 & 92.91+/-0.08 & 72.43+/-0.43 \\
        DELTA~\cite{li2019delta} & 88.35+/-0.41 & 94.76+/-0.05 &80.39+/-0.41 &  92.19+/-0.45 & 72.23+/-0.23 \\   
        BSS~\cite{chen2019catastrophic} & 88.34+/-0.62  & 94.84+/-0.21  &80.40+/-0.30 &  91.95+/-0.12 & 72.22+/-0.17 \\   
        %SSL-Contrast-CE-weighting & 66.46+/-0.61 &78.89+/-0.39  \\   
        RIFLE~\cite{li2020rifle}  & 89.06+/-0.28 &94.71+/-0.13 &  80.36+/-0.07 & 91.94+/-0.23  &72.45+/-0.30  \\
        SCL~\cite{gunel2020supervised}   & 89.29+/-0.07 & 95.33+/-0.09  & 81.49+/-0.27  &  92.84+/-0.03 & 72.73+/-0.31   \\    
        Bi-tuning~\cite{zhong2020bi}   &  89.06+/-0.08 &95.12+/-0.15 &   81.42+/-0.01 & 92.83+/-0.06  & 73.53+/-0.37  \\
        Core-tuning & \textbf{92.73+/-0.17} & \textbf{97.31+/-0.10} & \textbf{84.13+/-0.27} &  \textbf{93.46+/-0.06} & \textbf{75.37+/-0.37}   \\  
        \toprule
         
        \toprule
        Algorithm     &Aircraft& Cars & Pets & Flowers & Avg.  \\\toprule   
        SL-CE-tuning    &87.03+/-0.02 &  89.77+/-0.11 &  92.17+/-0.12 & 98.78+/-0.10 & 89.35  \\     \midrule
        CE-tuning  &86.87+/-0.18 &88.61+/-0.43 & 89.05+/-0.01 & 98.49+/-0.06  & 87.76  \\    
        L2SP~\cite{li2018explicit}  & 86.55+/-0.30  & 89.00+/-0.23 &  89.43+/-0.27 & 98.66+/-0.20 &  88.10 \\   
        M$\&$M~\cite{zhan2018mix}    &87.45+/-0.28 & 88.90+/-0.70 & 89.60+/-0.09  & 98.57+/-0.15 &  88.22\\
        DELTA~\cite{li2019delta}  &87.05+/-0.37  &88.73+/-0.05 &  89.54+/-0.48 & 98.65+/-0.17  & 87.99 \\   
        BSS~\cite{chen2019catastrophic} & 87.18+/-0.71  & 88.50+/-0.02 &  89.50+/-0.42  & 98.57+/-0.15   & 87.94\\   
        %SSL-Contrast-CE-weighting & 66.46+/-0.61 &78.89+/-0.39  \\   
        RIFLE~\cite{li2020rifle} & 87.60+/-0.50 &   89.72+/-0.11 & 90.05+/-0.26 & 98.70+/-0.06 & 88.29 \\
        SCL~\cite{gunel2020supervised}   &     87.44+/-0.31 & 89.37+/-0.13 &  89.71+/-0.20 &  98.65+/-0.10  & 88.54\\    
        Bi-tuning~\cite{zhong2020bi}   &87.39+/-0.01 &   89.41+/-0.28 & 89.90+/-0.06  & 98.57+/-0.10 & 88.58\\
        Core-tuning & \textbf{89.48+/-0.17}  & \textbf{90.17+/-0.03}   & \textbf{92.36+/-0.14}  & \textbf{99.18+/-0.15}  & \textbf{90.47} \\  
        \bottomrule
	\end{tabular}
    \end{threeparttable}}\vspace{-0.2in}
    \end{center}  
\end{table*} 

\begin{table*}[h]   
	\centering 
	\caption{Ablation studies of Core-tuning (Row 5) for fine-tuning MoCo-v2 pre-trained ResNet-50 on 9 natural image datasets in terms of top-1 accuracy. Here, ``Avg.'' indicates the average accuracy over the 9 datasets. Besides, $\mathcal{L}_{con}$ is the original supervised contrastive loss, while $\mathcal{L}^f_{con}$ is our focal contrastive loss. Moreover,  ``mix" denotes the manifold mix, while ``mix-H" indicates the proposed hardness-directed mixup   strategy in our method.}
	 %  \vspace{0.1in}
     \label{exp_ablation1} 
    \begin{center}
     \scalebox{0.75}{  
    \begin{threeparttable} 
	\begin{tabular}{c|cc|cc|ccccc}\hline
        $\mathcal{L}_{ce}$ & $\mathcal{L}_{con}$  &$\mathcal{L}^f_{con}$ & mix & mix-H &   ImageNet20 &CIFAR10 & CIFAR100 & Caltech101 & DTD    \\ \hline   
        %SL-CE & 94.23 & 83.40 & 93.65 & 74.40  &87.03 &  89.77 &  92.17 & 98.78  \\       
        $\surd$  &  &  & & & 88.28+/-0.47 & 94.70+/-0.39   &80.27+/-0.60  & 91.87+/-0.18  &   71.68+/-0.53  \\  
      $\surd$    &  $\surd$ & &  & & 89.29+/-0.07 & 95.33+/-0.09  & 81.49+/-0.27  &  92.84+/-0.03 & 72.73+/-0.31   \\  
        $\surd$  &  &  & $\surd$ & &90.67+/-0.09  & 95.43+/-0.20 &    81.03+/-0.11  & 92.68+/-0.06   &73.31+/-0.40    \\  

      $\surd$    & $\surd$  &  &  &$\surd$ & 92.20+/-0.15 & 97.01+/-0.10  & 83.89+/-0.20 &  93.22+/-0.18 &74.78+/-0.31    \\  
        $\surd$ &  &  $\surd$  &  &$\surd$ & \textbf{92.73+/-0.17}& \textbf{97.31+/-0.10} & \textbf{84.13+/-0.27} &  \textbf{93.46+/-0.06} & \textbf{75.37+/-0.37}    \\ 
 
        \hline
        \hline
        $\mathcal{L}_{ce}$ & $\mathcal{L}_{con}$  &$\mathcal{L}^f_{con}$ & mix &    mix-H & Aircraft& Cars & Pets & Flowers  & Avg. \\\hline  
        %SL-CE & 94.23 & 83.40 & 93.65 & 74.40  &87.03 &  89.77 &  92.17 & 98.78  \\       
        $\surd$  &  &  &  & &  86.87+/-0.18  &88.61+/-0.43 &89.05+/-0.01  & 98.49+/-0.06  & 87.76 \\  
      $\surd$    &  $\surd$  &  & &  &   87.44+/-0.31 & 89.37+/-0.13 &  89.71+/-0.20 &  98.65+/-0.10  & 88.54\\    
        $\surd$  &  &  & $\surd$  & &88.37+/-0.14  & 89.06+/-0.14 &91.37+/-0.03  & 98.74+/-0.11  & 88.96  \\

      $\surd$    & $\surd$  &  &  & $\surd$ &88.88+/-0.34  &89.79+/-0.12    & 91.95+/-0.33  & 98.94+/-0.12  & 90.07\\  
        $\surd$ &  &  $\surd$  &  &  $\surd$ & \textbf{89.48+/-0.17}  & \textbf{90.17+/-0.03}   & \textbf{92.36+/-0.14}  & \textbf{99.18+/-0.15} & \textbf{90.47} \\ 
 
        \hline
	\end{tabular} 
    \end{threeparttable}}
    \end{center}   %\vspace{0.1in}
     
\end{table*}

\textbf{The fine-tuning results on ImageNet.}  
Since ImageNet  has rich labeled samples for fine-tuning and the CSL models are also pre-trained on  ImageNet,  the performance gain of different fine-tuning methods may not vary as significantly as  on the small-scale target datasets. Even so, the results in Table~\ref{imagenet_result} also demonstrate the effectiveness of Core-tuning on very large-scale data. 
%In the main paper, we report the results of image classification and ablations studies on 9 natural image datasets in terms of the average accuracy.  To make the results more complete, this appendix further reports the results with their standard errors (cf.~Tables~\ref{imagenet_result}). 

	\begin{table}[h] 
		\caption{Fine-tuning results of the MoCo-v2  ResNet-50  fine-tuned by various methods, on ImageNet.}
     \label{imagenet_result} 	\vspace{-0.05in}
    \begin{center}
     \scalebox{0.85}{  
    \begin{threeparttable} 
	\begin{tabular}{llc}\toprule
	 Pre-training   &  Fine-tuning  &    Top-1 accuracy \cr
     \toprule 
      MoCo-v2~\cite{chen2020improved} &   CE-tuning  & 76.82   \\ 
      MoCo-v2~\cite{chen2020improved}  &   CE-Contrastive-tuning  & 77.13 \\  
      MoCo-v2~\cite{chen2020improved}  &   Core-tuning (ours)  & \textbf{77.43} \\  
        \bottomrule
	\end{tabular} 
  \end{threeparttable}}
    \end{center} % \vskip -0.05in  
	\end{table}	

\clearpage

\textbf{More results on different pre-training methods.} 
This appendix provides the fine-tuning results of Core-tuning for the SimCLR pre-trained models. Since the official checkpoints of SimCLR-v1~\cite{chen2020simple} and  SimCLR-v2~\cite{chen2020big} are based on Tensorflow, we convert them to the PyTorch  and try to  reproduce   cross-entropy tuning (CE-tuning) in our experimental settings. Note that although the reproduction performance of CE-tuning is slightly worse than the original paper~\cite{chen2020simple,chen2020big}, the results in this paper are obtained with the same preprocessing method \wrt each dataset and thus are fair. As shown  in Table~\ref{exp_ssl_model1}, Core-tuning consistently outperforms CE-tuning  for SimCLR pre-trained models.

% however, we find it difficult to reproduce the fine-tuning performance of some CSL models~\cite{chen2020simple}. Therefore,  for some datasets (\eg CIFAR10 and Aircraft), we try to resize images to different scales and use rotation augmentations. Although the preprocessing of some datasets is slightly different from~\cite{chen2020simple}, the results in this paper are obtained with the same preprocessing method \wrt each dataset and thus are more fair.

% , As shown  in Tables~\ref{exp_ssl_model1}. 

	\begin{table}[h] 
	\caption{Fine-tuning results of ResNet-50, pre-trained by various methods.}
     \label{exp_ssl_model1} 	\vspace{-0.05in}
    \begin{center}
     \scalebox{0.8}{  
    \begin{threeparttable} 
	\begin{tabular}{lcccccccc}\toprule
	 \multirow{2}{*}{Pre-training}        &   \multicolumn{2}{c}{Caltech101} &&  \multicolumn{2}{c}{DTD} &&  \multicolumn{2}{c}{Pets} \cr  \cmidrule{2-3}    \cmidrule{5-6} \cmidrule{8-9}  
     &  CE-tuning   & ours && CE-tuning   & ours  && CE-tuning   & ours \cr
     \toprule 
      SimCLR-v1~\cite{chen2020simple}  & 90.53+/-0.06 &\textbf{92.40+/-0.06}  &&90.53+/-0.06 & \textbf{71.26+/-0.05}  && 89.34+/-0.46 & \textbf{90.89+/-0.09} \\ 
     SimCLR-v2~\cite{chen2020big}   &  92.44+/-0.18 & \textbf{93.46+/-0.02}   && 71.26+/-0.26 & \textbf{74.75+/-0.41}  && 88.28+/-0.26 & \textbf{90.64+/-0.31}\\  
        \bottomrule
	\end{tabular} 
  \end{threeparttable}}
    \end{center}  %  \vskip -0.05in  
	\end{table}

\textbf{The results on linear evaluation.} 
This appendix provides linear evaluation for Core-tuning. Specifically, we first fine-tune the MoCo-v2 pre-trained ResNet-50 with Core-tuning and then train a linear classifier for  prediction. As shown  in Table~\ref{linear}, Core-tuning  performs better than CE-tuning.

	\begin{table}[h] 
		\caption{Results of linear evaluation for the ResNet-50  fine-tuned by various methods, on CIFAR10.}
     \label{linear} 	\vspace{-0.05in}
    \begin{center}
     \scalebox{0.85}{  
    \begin{threeparttable} 
	\begin{tabular}{llc}\toprule
	 Pre-training   &  Fine-tuning  &    Top-1 accuracy \cr
     \toprule 
      MoCo-v2~\cite{chen2020improved} &   CE-tuning  & 94.78+/-0.28   \\ 
      MoCo-v2~\cite{chen2020improved}  &   Core-tuning (ours)  & \textbf{97.09+/-0.14} \\  
        \bottomrule
	\end{tabular} 
  \end{threeparttable}}
    \end{center} % \vskip -0.05in  
	\end{table}	
	
\textbf{The results on KNN evaluation.} 
This appendix provides the KNN evaluation for Core-tuning. To be specific, we first fine-tune the MoCo-v2 pre-trained ResNet-50 with Core-tuning and then use KNN for prediction. As shown  in Table~\ref{KNN}, Core-tuning also outperforms CE-tuning.

	\begin{table}[h] 
	\caption{Results of KNN evaluation for the ResNet-50  fine-tuned by various methods, on CIFAR10.}
     \label{KNN} 	\vspace{-0.05in}
    \begin{center}
     \scalebox{0.85}{  
    \begin{threeparttable} 
	\begin{tabular}{llc}\toprule
	 Pre-training   &  Fine-tuning  &    Top-1 accuracy \cr
     \toprule 
      MoCo-v2~\cite{chen2020improved} &   CE-tuning  & 94.63+/-0.32   \\ 
      MoCo-v2~\cite{chen2020improved}  &   Core-tuning (ours)  & \textbf{96.65+/-0.06} \\  
        \bottomrule
	\end{tabular} 
  \end{threeparttable}}
    \end{center}  %\vskip -0.05in  
	\end{table}		
 
\subsection{The Results with Standard Errors on Semantic Segmentation}   
In the main paper, we report the average results of semantic segmentation on PASCAL VOC. This appendix further reports the results with their standard errors (cf.~Table~\ref{exp_segmentation1}). 

\begin{table}[H]    
 %\vskip -0.2in 
	\caption{Fine-tuning performance on PASCAL VOC semantic segmentation based on DeepLab-V3 with ResNet-50, pre-trained by various CSL methods.  CE indicates cross-entropy.}  
     \label{exp_segmentation1} 
    \begin{center}
    \scalebox{0.8}{  
    \begin{threeparttable} 
	\begin{tabular}{lcccc}\toprule
        Pre-training  & Fine-tuning&  MPA & FWIoU  & MIoU \\ \toprule     
        Supervised  &  CE & 87.10+/-0.20 & 89.12+/-0.17 & 76.52+/-0.34  \\  
        \midrule    
        \multirow{2}{*}{InsDis~\cite{wu2018unsupervised}}  & CE & 83.64+/-0.12 & 88.23+/-0.08  & 74.14+/-0.21 \\    
          & ours &  \textbf{84.53+/-0.31}  & \textbf{88.67+/-0.07} & \textbf{74.81+/-0.13}\\    
          %\hline
         \midrule    
         
        \multirow{2}{*}{PIRL~\cite{misra2020self}}  & CE & 83.16+/-0.26 & 88.22+/-0.24  & 73.99+/-0.42\\    
          & ours & \textbf{85.30+/-0.24}  & \textbf{88.95+/-0.08} &\textbf{75.49+/-0.36}      \\   \midrule   
      
          \multirow{2}{*}{MoCo-v1~\cite{he2020momentum}} & CE & 84.71+/-0.56 & 88.75+/-0.04 & 74.94+/-0.12\\    
          & ours & \textbf{85.70+/-0.32}  &\textbf{89.19+/-0.02}  & \textbf{75.94+/-0.23} \\    
          \midrule   
      
    \multirow{2}{*}{MoCo-v2~\cite{chen2020improved}}  & CE & 87.31+/-0.31 & 90.26+/-0.12 & 78.42+/-0.28\\    
          & ours & \textbf{88.76+/-0.34}  &\textbf{90.75+/-0.04}  & \textbf{79.62+/-0.12} \\    
          \midrule    
      
        \multirow{2}{*}{SimCLR-v2~\cite{chen2020big}} & CE  & 87.37+/-0.48 & 90.27+/-0.12  & 78.16+/-0.19 \\    
          & ours & \textbf{87.95+/-0.20}  & \textbf{90.71+/-0.13}  & \textbf{79.15+/-0.33} \\    \midrule   
      
      \multirow{2}{*}{InfoMin~\cite{tian2020makes}}  & CE & 87.17+/-0.20 & 89.84+/-0.09  & 77.84+/-0.24 \\    
          & ours & \textbf{88.92+/-0.36}  &\textbf{90.65+/-0.09}  & \textbf{79.48+/-0.30}      \\  
 
        \bottomrule
	\end{tabular}
	   \end{threeparttable}}
    \end{center}  % \vskip -0.2in   
\end{table} 
% InsDis~\cite{wu2018unsupervised}, PIRL~\cite{misra2020self}, MoCo-v1~\cite{he2020momentum} and InfoMin~\cite{tian2020makes}), clustering self-supervised methods (\ie SwAV~\cite{caron2020unsupervised} and DeepCluster-v2~\cite{caron2018deep})
\clearpage
\section{More Analysis of Core-tuning}      
\subsection{Analysis of Projection Dimension and Depth} 
\begin{figure}[h] 
 \begin{minipage}{0.47\linewidth}
 \centerline{\includegraphics[width=6cm]{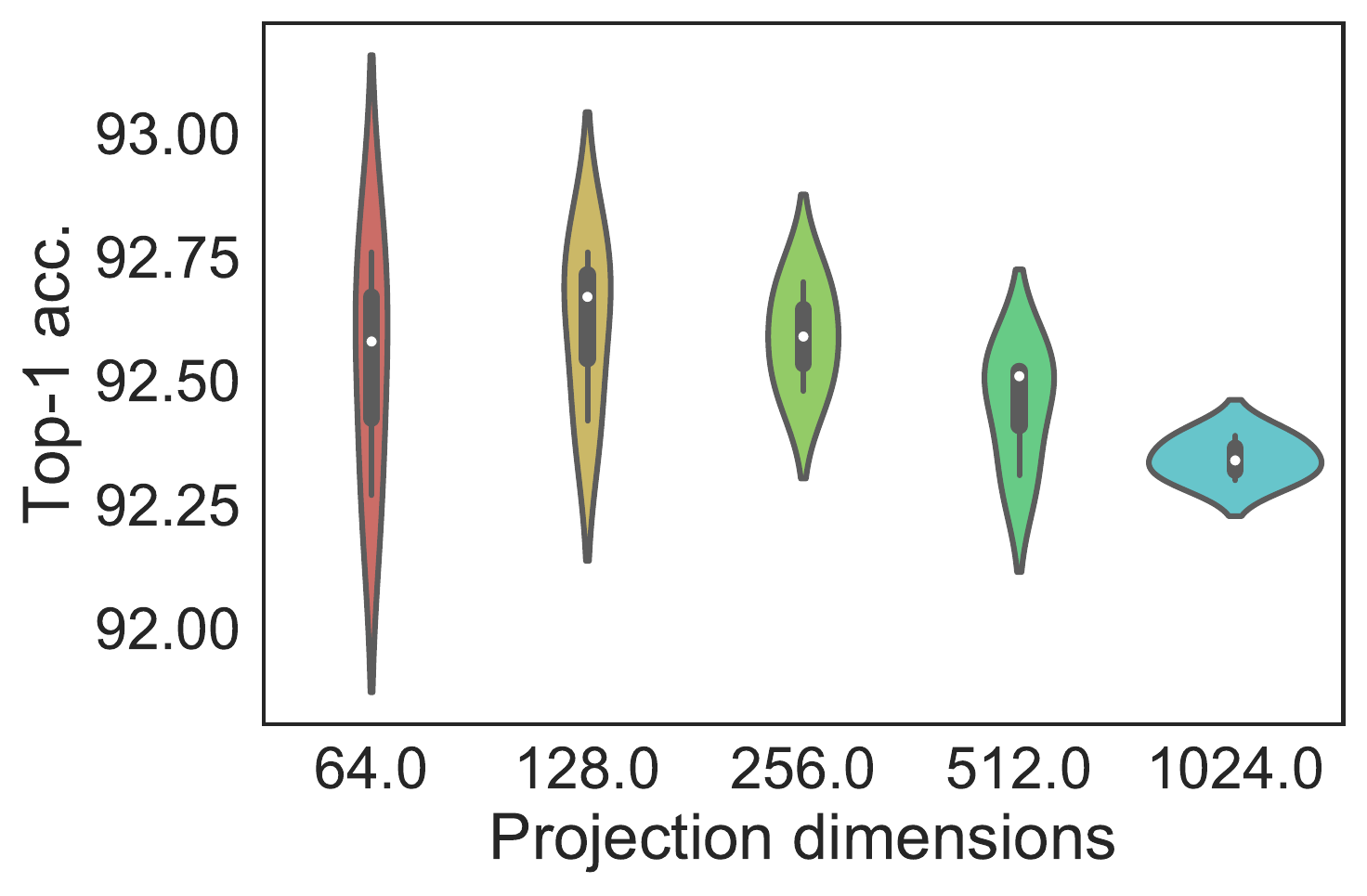}} 
 \end{minipage}
 \hfill 
 \begin{minipage}{0.47\linewidth}
 \centerline{\includegraphics[width=6cm]{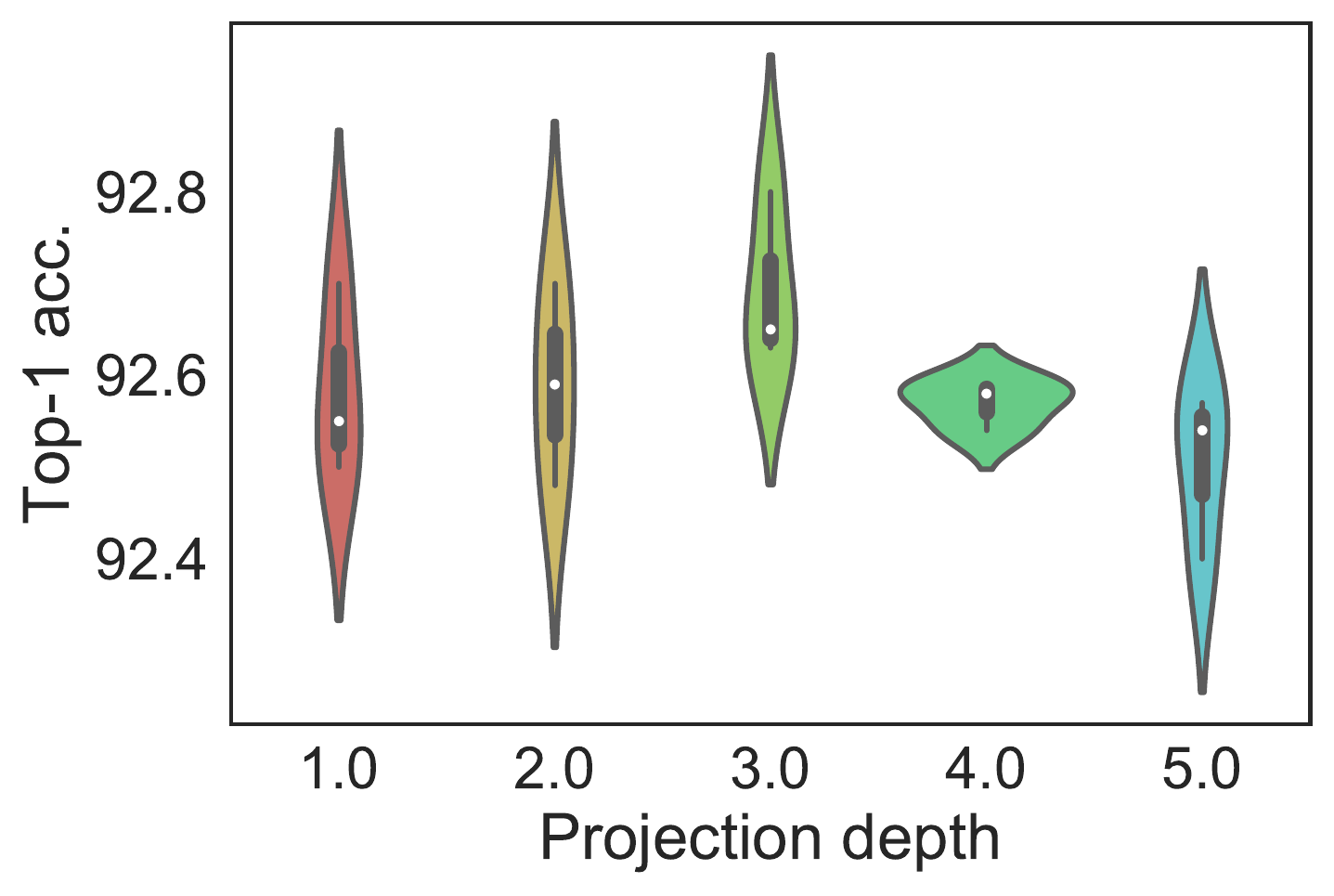}} 
  \end{minipage}  
 \caption{Analysis of the projection dimension  and the projection  depth in Core-tuning on  ImageNet20 based on MoCo-v2 pre-trained ResNet-50. Each run tests one parameter and fixes others. Best viewed in color.}
 \label{projection}  
\end{figure}

In previous experiments, we use a 2-layer MLP to extract contrastive features with dimension 256. Here, we further analyze how the dimension and the depth influence Core-tuning. The results  on ImageNet20 are reported in Figure~\ref{projection}, where the fine-tuning performance of Core-tuning can be further improved by changing the feature dimension to 128 and the depth to 3. Note that the best dimension and depth of the projection head may vary  on different datasets, but the  default setting (\ie dimension 256 and depth 2) is enough to obtain consistently  good performance. 

\subsection{Analysis of Loss and Mixup Hyper-Parameters}
This appendix discusses the influence of the loss trade-off parameter $\eta$ and the mixup sampling factor $\alpha$ on Core-tuning based  on the ImageNet20 dataset. Each run tests one parameter and fixes others. As shown  in Figure~\ref{parameter}, when $\eta\small{=}0.1$ and $\alpha\small{=}1$, Core-tuning performs slightly better on ImageNet20. Note that the best $\eta$ and $\alpha$ can be different on diverse datasets.

\subsection{Analysis of Temperature Factor}  
\begin{figure}[h]
 \begin{minipage}{0.31\linewidth}
 \centerline{\includegraphics[width=4.3cm]{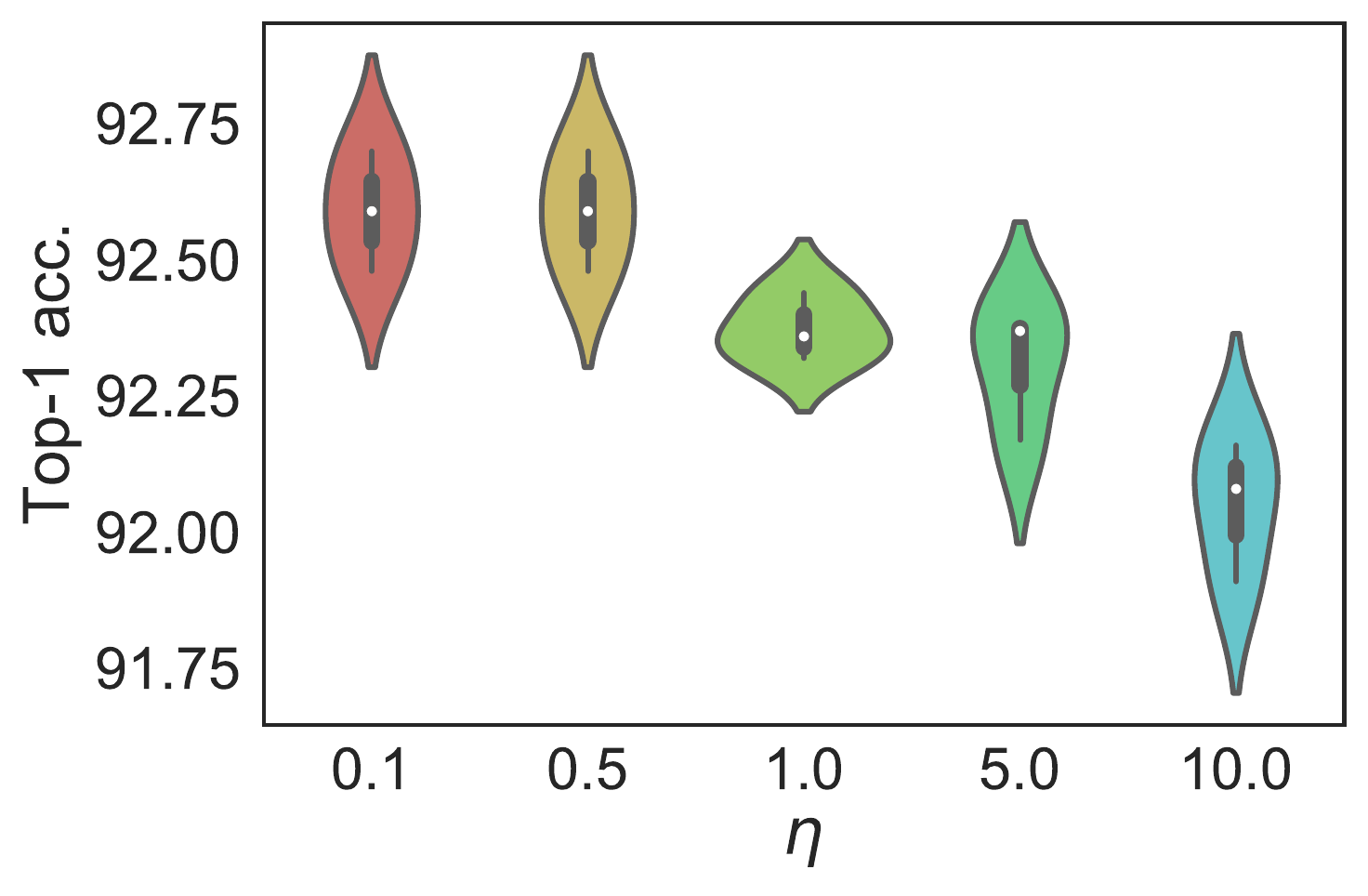}} 
 \end{minipage}
 \hfill 
 \begin{minipage}{0.31\linewidth}
\hspace{-0.2in} \centerline{\includegraphics[width=4.3cm]{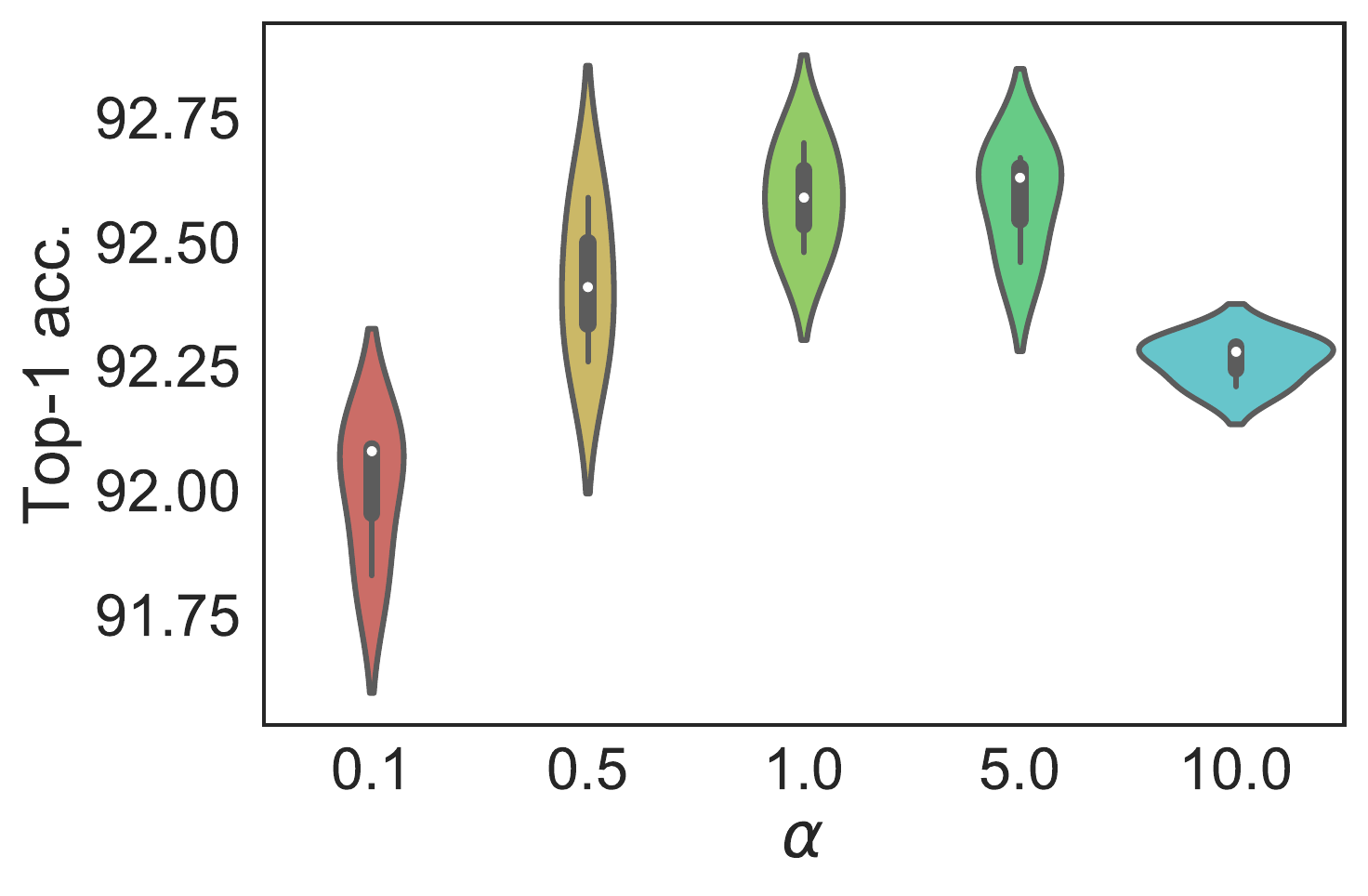}} 
  \end{minipage}
  \begin{minipage}{0.31\linewidth}
 \centerline{\includegraphics[width=4.3cm]{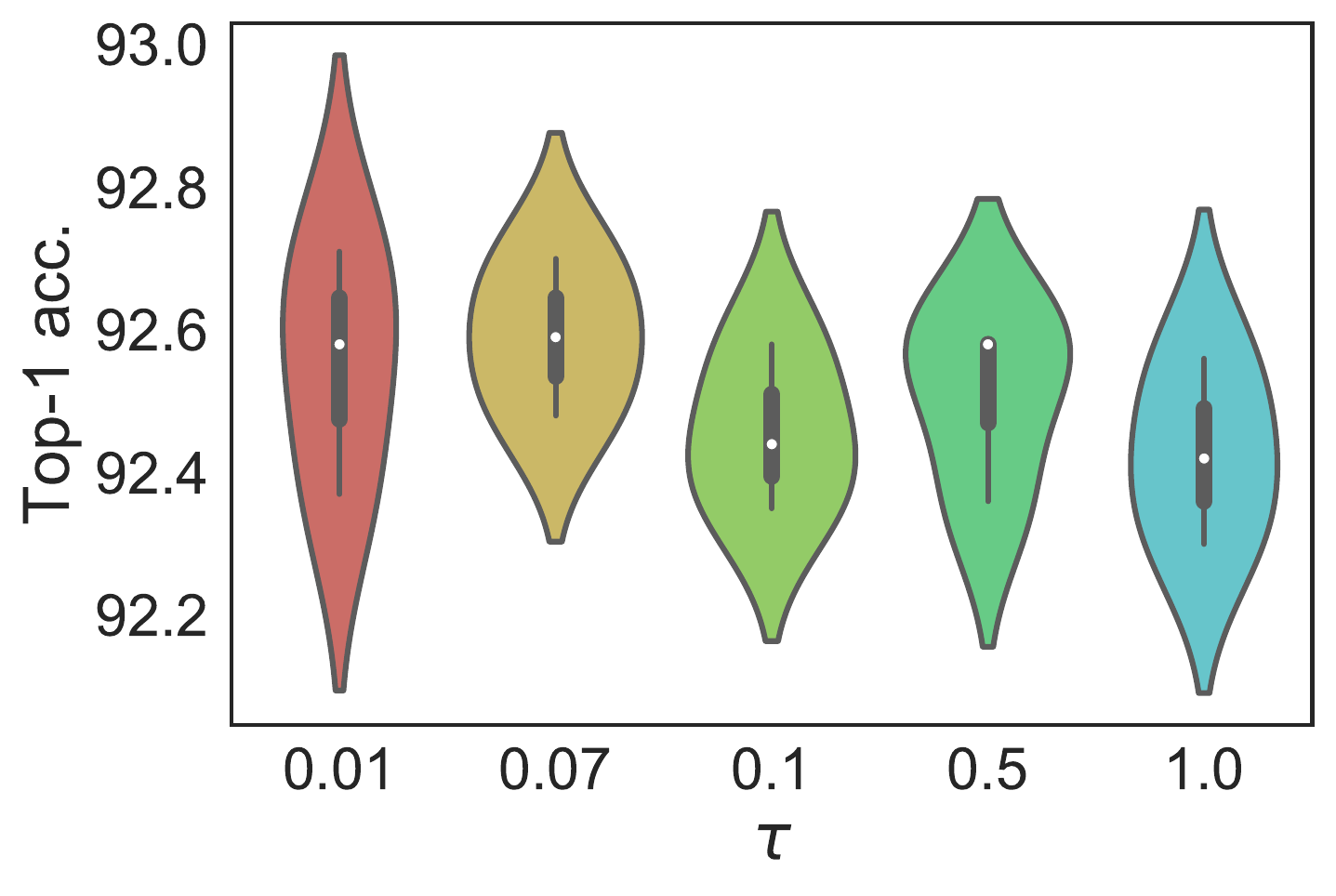}} 
  \end{minipage}
 \caption{Analysis of $\eta$, $\alpha$  and  the  temperature factor  in Core-tuning on  ImageNet20  based on MoCo-v2 pre-trained ResNet-50. Each run tests one factor and fixes others. Best viewed in color.}
 \label{parameter}  
\end{figure}

Following the implementation of the supervised contrastive loss~\cite{khosla2020supervised}, we set the temperature factor $\tau$ to $0.07$ for Core-tuning by default. In this section, we further analyze the influence of  $\tau$ on Core-tuning when fine-tuning MoCo-v2 pre-trained models on ImageNet20. As shown in Figure~\ref{parameter},  when $\tau$ is small (\eg 0.01 or 0.07), Core-tuning performs slightly  better on ImageNet20. The potential reason is that a small temperature parameter implicitly helps the method to learn hard positive/negative pairs~\cite{wang2020understanding1}, which are more informative and beneficial to contrastive learning.   Note that the best $\tau$ can be different on different datasets, but   the  default setting (\ie $\tau=0.07$) is enough to achieve comparable performance.

% To make the generated negative pairs closer to negatives, we clip $\lambda\small{\sim}\text{Beta}(\alpha,\alpha)$ by $\lambda \small{\geq}\lambda_n$ when mixing hard negative pairs, where $\lambda_{n}$ is a threshold and we set it to 0.8.  
  
\clearpage
\subsection{Analysis of Hard Pair Thresholds} 
In our hardness-directed mixup strategy, to make the generated negative pairs closer to negative pairs, we  clip $\lambda\small{\sim}\text{Beta}(\alpha,\alpha)$ by $\lambda \small{\geq}\lambda_n$ when generating hard negative pairs. In our experiments, we set the threshold $\lambda_{n}=0.8$. In this appendix, we analyze the influences of the  negative pair threshold $\lambda_n$. Meanwhile, although we do not constrain hard positive generation, we also analyze  the potential positive pair threshold $\lambda_p$.  The results  on ImageNet20  are reported in Table~\ref{parameter}. On the one hand, $\lambda_n$ satisfies our expectation that the generated hard negative pairs should be closer to negatives, \ie a larger $\lambda_n$ can lead to better performance. On the other hand,  we find  when no crop  is conducted for hard positive generation (\ie $\lambda_p\small{=}0$), the performance is slightly better. We conjecture that since the generated hard positives are  located in the borderline area between positives and negatives, allowing the generated hard positives to close to negatives may have a margin effect on contrastive learning and thus boosts performance. Despite this, Core-tuning with a large $\lambda_p$    performs similarly well.

\begin{table}[h]    
 %\vskip -0.1in 
	\caption{Threshold analysis for hard pair generation  in  Core-tuning on  ImageNet20 based on    MoCo-v2 pre-trained  ResNet-50. Each run tests one parameter and fixes another one  to 0.8.}
     \label{exp_threshold} 
    \begin{center}
    \scalebox{0.9}{  
    \begin{threeparttable} 
	\begin{tabular}{lccccc}\toprule
        Thresholds  & 0&  0.2 & 0.4  & 0.6  & 0.8 \\\toprule      
        Negative pair threshold $\lambda_n$ & 91.55 & 91.94  &92.19  & 92.36 &92.59\\   
        Positive pair threshold  $\lambda_p$   & 92.73 & 92.68 & 92.64 & 92.60 & 92.59 \\     
        \bottomrule
	\end{tabular}
	   \end{threeparttable}}
    \end{center}  % \vskip -0.2in 
\end{table}

% \subsection{\blue{Analysis of Only Using Hard Negative Pairs for Classifier Training}} 
% In the main paper, we report the results of image classification and ablations studies on 9 natural image datasets in terms of the average accuracy.  To make the results more complete, this appendix further reports the results with their standard errors, as shown  in Tables~\ref{exp_hard_negative}. 

% 	\begin{table}[h] 
% 		\caption{Comparisons with  only using  hard negative pairs for contrastive fine-tuning on CIFAR10.}
%      \label{exp_hard_negative} 	\vspace{-0.05in}
%     \begin{center}
%      \scalebox{0.9}{  
%     \begin{threeparttable} 
% 	\begin{tabular}{llc}\toprule
% 	 Pre-training   &  Fine-tuning  &    Top-1 accuracy \cr
%      \toprule 
%       MoCo-v2~\cite{chen2020improved} &   Core-tuning (only hard negative) &  97.31+/-0.10   \\ 
%       MoCo-v2~\cite{chen2020improved}  &   Core-tuning (full)  &  97.34+/-0.07 \\  
%         \bottomrule
% 	\end{tabular} 
%   \end{threeparttable}}
%     \end{center}  %\vskip -0.1in  
% 	\end{table}	

% \newpage
\subsection{Relationship Between Pre-Training and Fine-Tuning Accuracies} 
\begin{figure}[h]   
 \centerline{\includegraphics[width=8cm]{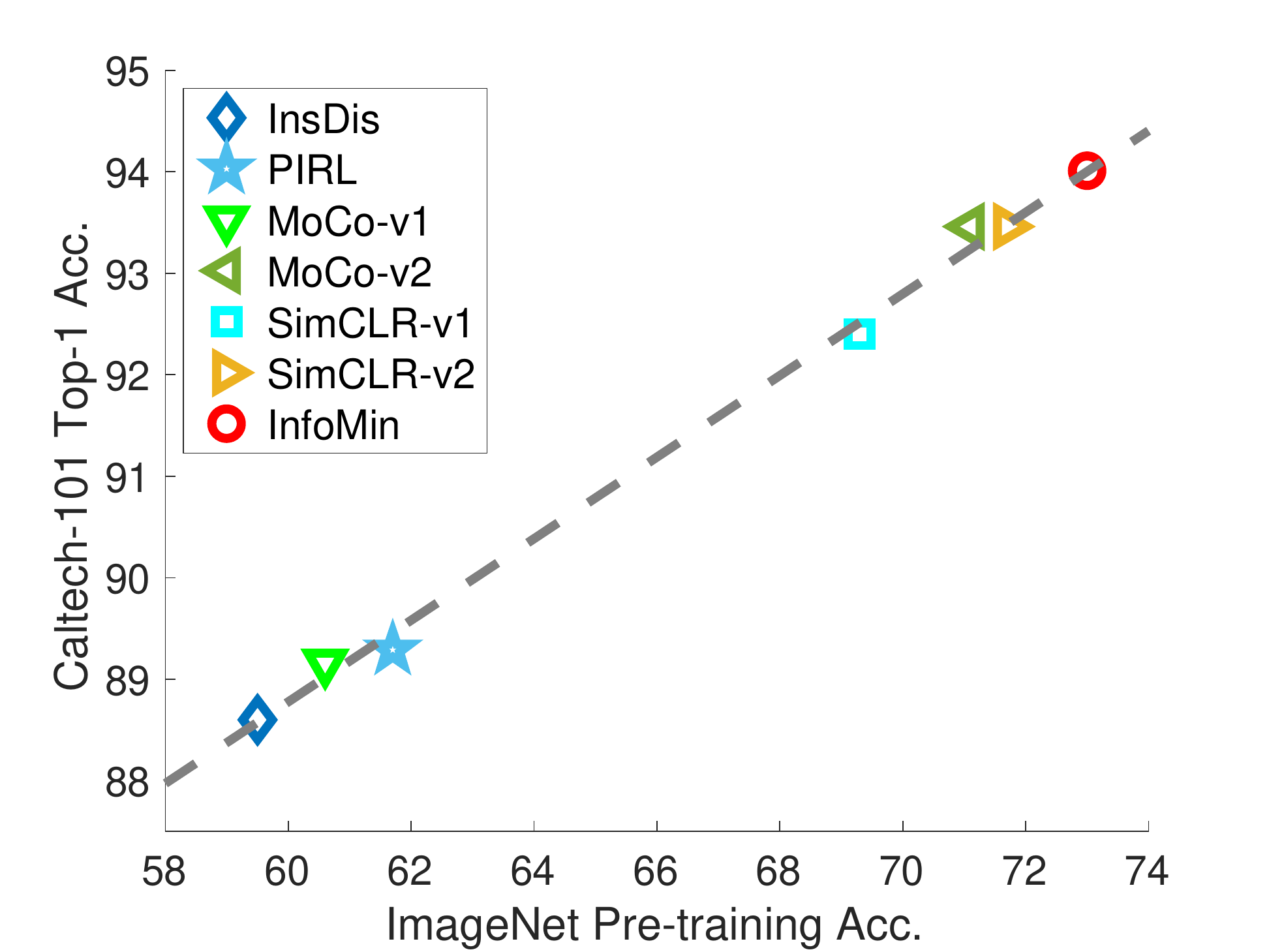}} %\vskip -0.1in
 \caption{The relationship between ImageNet performance and Core-tuning fine-tuning performance   on Caltech-101 for  contrastive self-supervised ResNet-50 models. Better viewed in color.}
 \label{correlation}  
\end{figure}

We further explore the relationship between ImageNet performance and  Core-tuning fine-tuning performance  on Caltech-101 for various contrastive self-supervised models. Here, the ImageNet performance of a contrastive self-supervised model  is obtained by training a new linear classifier on the frozen pre-trained representation and then evaluate the model on the ImageNet test set. For convenience, we directly follow the ImageNet performance reported in the original paper of the corresponding methods. 
As shown in Figure~\ref{correlation}, the fine-tuning result of each contrastive self-supervised model  on Caltech-101 is highly correlated with the  model result on ImageNet. This implies that the ImageNet performance can be a good predictor for the fine-tuning performance of contrastive self-supervised models. Such a finding is consistent with supervised pre-trained models~\cite{kornblith2019better}.
Even so, note that the correlation is not perfect, where a  contrastive pre-trained model with better ImageNet performance does not necessarily mean better fine-tuning performance, \eg SimCLR-v2 vs MoCo-v2. 

\subsection{Effectiveness of Hard   Pair Generation for Contrastive Fine-Tuning} 
In our proposed Core-tuning, we use all the generated positive sample pairs and the original samples as positive pairs for  contrastive fine-tuning. In this appendix, to better evaluate the effectiveness of hard pair generation, we do not use  original data as positive pairs but only use the generated hard positive pairs for contrastive learning. As shown  in Table~\ref{exp_hard_positive}, only using the generated hard positive pairs for contrastive learning is enough to obtain comparable performance. Such results further verify the effectiveness of our hardness-directed mixup strategy as well as the importance of hard positive pairs for contrastive fine-tuning.

	\begin{table}[h] 
		\caption{Comparisons with  only using the generating hard positive pairs for contrast on CIFAR10.}
     \label{exp_hard_positive}  
    \begin{center}
     \scalebox{0.95}{  
    \begin{threeparttable} 
	\begin{tabular}{cccc}\toprule
	 Pre-training   &  Fine-tuning  & The used positive pairs for contrast? &  Top-1 accuracy \cr
     \toprule 
     MoCo-v2~\cite{chen2020improved}  &   CE-tuning   &  $\times$  & 94.70+/-0.39 \\
      MoCo-v2~\cite{chen2020improved} &   Core-tuning & only the generated hard positive pairs &   97.31+/-0.09   \\ 
      MoCo-v2~\cite{chen2020improved}  &   Core-tuning & all positive pairs &  97.31+/-0.10  \\  
        \bottomrule
	\end{tabular} 
  \end{threeparttable}}
    \end{center}  %\vskip -0.1in  
	\end{table}

\subsection{Effectiveness of Smooth Classifier Learning} 
In Core-tuning, to better exploit the learned discriminative feature space by contrastive fine-tuning, we use the mixed samples for classifier training, so that the classifier can be more smooth and far away from the original training data. In this appendix, to better evaluate the effectiveness of smooth classifier learning, we compare Core-tuning with a variant that does not use the mixed data for classifier learning. As shown  in Table~\ref{exp_Smooth}, smooth classifier learning contributes to the fine-tuning performance of contrastive self-supervised models on downstream tasks. The results demonstrate the effectiveness of smooth classifier learning and also show its importance in  Core-tuning.

	\begin{table}[h] 
		\caption{Influence of smooth classifier learning  on CIFAR10.}
     \label{exp_Smooth} 	 
    \begin{center}
     \scalebox{0.95}{  
    \begin{threeparttable} 
	\begin{tabular}{cccc}\toprule
	 Pre-training   &  Fine-tuning  &  Smooth classifier learning? &  Top-1 accuracy \cr
     \toprule 
      MoCo-v2~\cite{chen2020improved} &   CE-tuning & $\times$  &  94.70+/-0.39 \\
      MoCo-v2~\cite{chen2020improved} &   CE-tuning & $\surd$   & 95.43+/-0.20\\
      MoCo-v2~\cite{chen2020improved} &   Core-tuning & $\times$  &  96.13+/-0.11 \\ 
      MoCo-v2~\cite{chen2020improved}  &   Core-tuning & $\surd$  &  97.31+/-0.10  \\  
        \bottomrule
	\end{tabular} 
  \end{threeparttable}}
    \end{center}  %\vskip -0.1in  
	\end{table}

\end{document}